\documentclass{article}

\PassOptionsToPackage{numbers, compress}{natbib}

\usepackage[preprint]{neurips_2026}

\usepackage{svg}
\usepackage[utf8]{inputenc}
\usepackage[T1]{fontenc}
\usepackage{hyperref}
\usepackage{url}
\usepackage{booktabs}
\usepackage{amsfonts}
\usepackage{amsmath}
\usepackage{empheq}
\usepackage{algorithm}
\usepackage{algpseudocode}
\usepackage{amsthm}
\DeclareMathOperator{\sign}{sign}
\DeclareMathOperator*{\argmax}{arg\,max}

\DeclareMathOperator{\KL}{KL}

\usepackage{amssymb}
\newcommand{\Xset}{\mathcal{X}}
\usepackage{nicefrac}
\usepackage{microtype}
\usepackage{xcolor}
\usepackage{graphicx}
\usepackage{soul}
\usepackage{comment}
\usepackage{multirow}

\usepackage{tikz}
\usetikzlibrary{positioning, arrows.meta, shapes.geometric, fit, calc}

\definecolor{navy}{HTML}{1F3A5F}
\definecolor{teal}{HTML}{4A8B8C}
\definecolor{warmgray}{HTML}{8B8B8B}
\definecolor{accent}{HTML}{E8946A}
\usepackage{wrapfig}

\newcommand{\vg}[3]{#1 (#2\,/\,#3)}

\usepackage{tcolorbox}
\tcbuselibrary{breakable, skins}

\newtcolorbox{promptbox}[1][]{%
  colback=gray!5,
  colframe=gray!50,
  fonttitle=\bfseries\small,
  title={#1},
  breakable,
  enhanced,
  left=6pt, right=6pt,
  top=4pt, bottom=4pt
}

\title{Zero-Shot Active Feature Acquisition via LLM-Elicitation.}

\author{%
  Binyamin Perets*\\
  Faculty of EE, Technion. \\
  Faculty of Medicine, Technion. \\
  \And
  Natalie Mendelson*\\
  Faculty of EE, Technion. \\
  Faculty of Medicine, Technion. \\
  \AND
  Shiran Vainberg \\
  CytoReason \\
  \And
  Yehuda Chowers \\
  CytoReason \\
  \AND
  Shai S. Shen-Orr \\
  Faculty of Medicine, Technion. \\
  \And
  Shie Mannor \\
  Faculty of EE, Technion. \\
  NVIDIA \\
}

\begin{document}

\maketitle

\begin{abstract}
Active feature acquisition (AFA) sequentially selects which features to observe to reach a classification or ranking decision. Its central limitation is reliance on large amount of labeled data to fit probabilistic models guiding acquisition. Large language models (LLMs) supply unsupervised domain knowledge, but are poor sequential planners. Asking one to both know and decide conflates capabilities best kept separate.
Here, we develop a framework for zero-shot AFA through disciplined elicitation: asking the LLM only for what it can be trusted to return, the unary deviations and pairwise co-variations that are the sufficient statistics of a Markov random field (MRF). We apply our framework to two settings: binary classification and top-$k$ identification. In practice, the LLM reliably returns only discriminative statistics, what distinguishes the classes rather than each class in isolation, which precludes classical AFA. We apply a maximum-entropy closure that resolves this gauge ambiguity.
We evaluate on a cohort of Inflammatory Bowel Disease (IBD) patients, an active clinical setting where diagnostic ambiguity and patient heterogeneity obstruct stable treatment strategies. Our framework outperforms the LLM both on real labels and on its own extracted beliefs. Where it matters most, on the hardest patients, our top-$k$ acquisition policy markedly outperforms all existing methods.
\end{abstract}

\section{Introduction}
\label{sec:intro}

Many real-world decisions require sequential, adaptive measurement under a limited budget, sharing the same structure: observe, update, choose what to observe next. And in each, measurement cost and the combinatorial space of orderings render exhaustive strategies infeasible. This defines \emph{Active Feature Acquisition (AFA)}: a sequential decision task in which the ordering of observations matters, and optimal ordering requires a model of how features relate to hypotheses. AFA relies on constructing a probabilistic model that encodes which measurements are informative for which hypotheses, used downstream to guide the acquisition process. Typically, this model is learned from large labeled datasets in which both features and ground-truth labels are known \citep{Ma2019, Shim2018, saartsechansky2009active}. This requirement is the fundamental bottleneck: labeled datasets are expensive to collect, highly domain-specific, and in multi-label settings, where multiple hypotheses may be simultaneously active, extraordinarily difficult to obtain at scale \citep{Janisch2019, Covert2023}. This scarcity is sharpest in precisely the regimes where AFA is most valuable: rare or newly-characterized diseases for which training data is limited or unavailable, and atypical patients whose classification requires many features rather than a few discriminative ones. \textit{Zero-shot AFA}, operating with no task-specific labeled
data, is therefore both necessary and largely unsolved; a recent
survey of AFA methods~\citep{rahbar2025survey} confirms that every
existing approach requires labeled data, RL episodes, or a
meta-training distribution.

Large language models (LLMs) are a natural candidate to break this bottleneck. Trained on the entirety of the scientific and medical corpus, SOTA models have shown strong potential to generalize to unseen domains and labels \cite{Brown2020, Wei2022, Bubeck2023, Kojima2022}, precisely the regime where labeled data is unavailable. Crucially, SOTA proprietary models substantially outperform open alternatives in knowledge quality, calibration, and structured-output reliability \cite{OpenAI2023, Anthropic2024, Zheng2023}, making them the natural knowledge source for zero-shot model construction. This promise, however, collides with two fundamental obstacles. First, AFA requires sequential optimization over an exponential space of future observations, a task at which LLMs are demonstrably poor. They pattern-match heuristically, preferring features that sound important over features that are discriminatively optimal, and fail to reason about the cascading, non-local implications of feature correlations in a graphical model \cite{Valmeekam2023, Kambhampati2024, Stechly2024}. Second, AFA planning requires a white-box model one can condition, marginalize, and optimize over, but LLMs are poorly calibrated as formal probabilistic objects \cite{Kadavath2022, Zhao2021}. Combined with the label scarcity above, this leaves human expert interrogation as the de facto fallback in exactly the settings where automated AFA is most needed. As we also show empirically, even when the LLM is given its own extracted model in context and asked to plan over it, it fails significantly, following the results of recent benchmarks \cite{Gupta2025}.
In this work we aim to bridge this gap, allowing for a principled route to zero-shot AFA grounded in LLM knowledge. Our path is through a strict separation of roles: the LLM populates a formal probabilistic model offline; a classical algorithm plans over it online. Under this separation, we extract from the LLM only quantities that are well-posed for it: the unary marginal (``under hypothesis $H$, does feature $F$ deviate from baseline, and if so in which direction?''), and the pairwise relations (``under hypothesis $H$, to what extent do features $F$ and $G$ co-vary?''). These are exactly the sufficient statistics of an MRF. 

A central design choice in our work, and a departure from standard AFA pipelines, is to avoid ever explicitly representing the two class-conditional distributions $p(\mathbf{x} \mid H)$ and $p(\mathbf{x} \mid \neg H)$. While one could in principle elicit two MRFs, one asking ``what defines a sample under $H$'' and another asking ``what defines a sample not under $H$,'' and difference them post hoc, we argue this is the wrong primitive for two reasons. First, the structure of the world's knowledge, particularly in medicine, is discriminative rather than generative: the literature on a disease tells us how an inflamed colon's expression profile differs from a healthy colon's, not the absolute distribution of either. Second, eliciting distributions in absolute terms forces the LLM to commit to questions with no useful answer. We therefore extract from the LLM only discriminative information: how a feature deviates and co-varies under one hypothesis relative to another. This creates an identification problem: a log-ratio does not determine the two class-conditionals, since any pair inducing it is equally consistent, so the acquisition criteria of classical AFA, which depend on the per-class distributions (e.g., conditional mutual information), cannot be computed from the ratio alone. We resolve this with a maximum-entropy closure: the unique class-conditional pair consistent with the elicited log-ratio that introduces no information beyond it.
This work addresses  two settings: the classical binary classification, which is the common case in AFA literature, and the surprisingly understudied problem of top-$k$ identification: selecting the $k$ most supported hypotheses from a large candidate set, underlying many hallmark problems (e.g differential diagnosis). For binary classification, the standard criterion~\cite{Ma2019, Covert2023} is not trivially applied due to the discrimination of elicited information. This identification problem is the core methodological obstacle for the binary case and we show that under the MaxEnt gauge we can resolve it. As an alternative that sidesteps the gauge, we compare it to a Wald-like accumulator. Extending to top-k, we introduce a novel policy inspired by the dueling-bandit framework. Although our problem is not a classical bandit setting, we draw inspiration from the dueling-bandit literature for reasons rooted in MRF tractability. MRFs are hard to track exactly and admit only approximate inference, so reasoning over a \emph{difference} between two MRFs is structurally easier than reasoning over either one in isolation. Dueling also provides a clean notation for ordering. Concretely, we instantiate the framework as a \emph{preference score} that orders any two entities given the current observations, paired with a pairwise allocation rule that decides which duel to advance next.

Beyond the application, the framework enables a controlled measurement of the \emph{knowledge--planning gap} in LLMs. We compare our algorithm, which uses the LLM-derived MRF for formal planning, against an in-context LLM baseline given the same MRF and the same task. Knowledge is identical across both arms; only the planner differs. In this maximally generous setting, the formal algorithm reaches the same accuracy with much fewer observations, a gap attributable entirely to planning.

The remainder of the paper is organized as follows: \S\ref{sec:related} reviews related works and positions ours. \S\ref{sec:problem} formalizes the binary classification and top-$k$ identification settings. \S\ref{sec:solution} develops the solution: the MaxEnt closure \citep{jaynes1957information} that resolves the gauge ambiguity in the binary case (\S\ref{sec:landscape}), the allocation policy (\S\ref{sec:pref_scores}--\S\ref{sec:gain}), and a novel priority-weighted allocation rule for top-$k$ (\S\ref{sec:topk_alloc}). \S\ref{sec:experiments} evaluates the framework on a real IBD cohort, isolating the contribution of each component and quantifying the LLM knowledge--planning gap. \S\ref{sec:discussion} and \S\ref{sec:limitations} close with implications and open problems.

\subsection{Problem Setup}
\label{sec:problem}
We study active feature acquisition in two settings: binary classification and top-$k$ identification. In both settings, an entity $v$ admits two latent states $y \in \{0,1\}$ (baseline and condition), each inducing a class-conditional distribution over $M$ features $\mathbf{x} = (x_1, \dots, x_M)$, $x_j \in \{-1, 0, +1\}$, modeled as an MRF with unary and pairwise potentials:
\begin{equation}
  P(\mathbf{x} \mid y)
  = \frac{1}{Z^{(y)}}
  \exp\bigg(
    \sum_{j=1}^{M} \alpha_j^{(v,y)} x_j
    + \sum_{(j,l) \in \mathcal{E}} \psi_{jl}^{(v,y)} x_j x_l
  \bigg).
  \label{eq:class_mrf}
\end{equation}
Here $\alpha_j^{(v,y)} \in \mathbb{R}$ is the unary potential of feature $j$ under state $y$ for entity $v$, $\psi_{jl}^{(v,y)} \in \mathbb{R}$ is the pairwise potential coupling features $j$ and $l$, and $Z^{(y)}$ is the normalizing partition function. An MRF relies on the Markov property: for edge set $\mathcal{E} \subseteq \binom{[M]}{2}$, $x_j \perp\!\!\!\perp x_l \mid \mathbf{x}_{\setminus\{j,l\}}$ whenever $(j,l) \notin \mathcal{E}$. The use of MRFs relies on two observations: (1) Unlike similar models, the graph is symmetric and undirected, matching the symmetric form of LLM elicitation. (2) Unlike general factor graphs, interactions are restricted to pairs. 

As described before, throughout the paper we address the case where \textbf{the two class-conditionals in~\eqref{eq:class_mrf} are a conceptual model only, that is, we never have access to each separately.} Instead, LLM elicitation yields the contrast between hypotheses: the deviation of feature $j$ under $y=1$ relative to $y=0$, and their differential covariation. We formalize these as the parameters of the \emph{discriminative MRF}, with unaries $\Delta\alpha_j^{(v)} = \alpha_j^{(v,1)} - \alpha_j^{(v,0)}$ and pairwise couplings $\Delta\psi_{jl}^{(v)} = \psi_{jl}^{(v,1)} - \psi_{jl}^{(v,0)}$. Assuming equal priors, the discriminative score computes the log-ratio directly:
\begin{equation}
  S_v(\mathbf{x})
  = \sum_{j=1}^{M} \Delta\alpha_j^{(v)} x_j
  + \sum_{(j,l) \in \mathcal{E}} \Delta\psi_{jl}^{(v)} x_j x_l.
  \label{eq:disc_score}
\end{equation}
The discriminative MRF parameterized by these contrasts—not the per-class MRFs of~\eqref{eq:class_mrf}—is the model on which we operate.

\textbf{The acquisition protocol }follows: features are observed sequentially,that is, at round $t$ the feature set is partitioned into observed $\mathbf{x}_{B^{(t)}}$ and unobserved $\mathbf{x}_{A^{(t)}}$, with $B^{(0)} = \emptyset$. The algorithm selects which feature to observe next, receives its value (deterministic), and updates. For \textbf{binary classification}, a single entity $v$ is given and the task is to determine the true hypothesis using the fewest observations. For \textbf{top-$k$ identification}, $N$ entities $\{v_1, \dots, v_N\}$ are given, and the task is to identify the $k$ entities for which $y = 1$ is most supported, again using the fewest observations.

\section{Related Work}
\label{sec:related}

\textbf{Active feature acquisition.}
AFA originates in cost-sensitive classification \citep{melville2004active, Schutz2025}: a classifier $P(Y \mid \mathbf{x}_\mathrm{obs})$ over partially observed instances is maintained, and each candidate feature is scored by an acquisition criterion, typically the expected reduction in posterior entropy or the conditional mutual information (CMI) with the target. Evaluating any such criterion requires a predictive model $P(x_j \mid \mathbf{x}_\mathrm{obs})$, either learned explicitly or approximated through the classifier. AFA methods vary along two axes. \emph{Myopicity}: greedy policies select the feature with highest immediate expected utility, while non-myopic policies account for the downstream value of future observations. \emph{Model type}: generative methods learn $p(x_i \mid \mathbf{x}_\mathrm{obs})$ and plug it into the criterion, while discriminative methods parameterize the acquisition value directly. Myopic generative methods, such as EDDI~\citep{Ma2019} and ACFlow~\citep{li2021acflow}, estimate the conditional density and greedily maximize CMI. Myopic discriminative methods~\citep{Covert2023, dime2023, difa2024} estimate CMI directly over arbitrary observed subsets, recovering the Bayes-optimal greedy policy at convergence. Non-myopic methods target the value of future observations via RL~\citep{Shim2018, odin2023} or oracle lookahead (AACO~\citep{aaco2024}). Meta-learning approaches~\citep{l2m2024} amortize policies across task distributions. Our framework departs along three dimensions. First, every existing AFA method requires labeled data, RL episodes, or a meta-training distribution; ours requires none, parameterizing the MRF entirely from LLM queries. Second, existing models are opaque with respect to unobserved features, precluding deterministic bounding. Third, existing methods are binary, whereas we additionally rank $N$ entities sharing the same feature space.

\textbf{Sequential decision making.}
The formal study of sequential decision making originates with Wald's sequential probability ratio test (SPRT)~\citep{wald1945sequential}, developed for industrial quality control under costly inspection: at each round, the running log-likelihood ratio between two hypotheses is computed from the observations seen so far, and the test continues until the ratio crosses a decision threshold. SPRT is an optimal stopping rule over a fixed observation stream; it decides only when to halt and which hypothesis to accept, with no notion of which observation to take next. Chernoff~\cite{chernoff1959sequential} introduced the active sequential version with a heuristic randomized policy: the next observation is chosen to maximize the KL divergence between the two hypothesis distributions, the procedure AFA inherits. We adopt the Wald--Chernoff combination into our framework.

\textbf{LLMs and planning.}
A growing body of work shows that LLMs struggle with sequential planning even given complete problem specifications. \cite{Valmeekam2023} found GPT-4 achieves below 35\% on Blocksworld with full PDDL in context; PlanBench~\citep{valmeekam2023planbench} extended this across domains. \cite{Kambhampati2024} synthesized these into the LLM-Modulo framework, arguing LLMs are best viewed as approximate knowledge sources paired with external verifiers. Thought of Search~\citep{thought2024search} showed that extracting search components symbolically and performing search without LLM involvement achieves 100\% accuracy where LLM planners fail. In scientific domains, \cite{Gupta2025} found LLM-based Bayesian optimization agents insensitive to experimental feedback, and hybrid approaches (LLAMBO~\citep{llambo2024}, BORA~\citep{bora2025}) confirm that LLMs are most effective contributing knowledge to formal optimization frameworks.

\textbf{LLMs for probabilistic model construction.}
Recent work uses LLMs to parameterize graphical models: \cite{llm2025bn} showed LLM-derived CPTs for Bayesian networks outperform uninformative baselines in low-data regimes, and \cite{domke2025large} proposed generating formal probabilistic programs from LLMs for standard Bayesian inference. We share the principle of LLM-populated formal models but target active sequential acquisition on MRFs rather than static inference on Bayesian networks.

\section{The Zero-Shot AFA Framework}
\label{sec:solution}

The discriminative scaffold of \S\ref{sec:problem} reduces both settings to the same per-step primitive: score candidate features, select one, observe its value, update the diff-MRF state (App. \ref{supp:updatemrf}), recompute the running answer for the decision target. The two settings differ in what that answer is: binary classification asks for the sign of one discriminative score $S_v$, and top-$k$ asks for a partial ordering over $N$ entities, intuitively, the entities for which the positive-label hypothesis dominates the negative through the log-odds of the discriminative MRF. Because top-$k$ is an ordering problem, the natural primitive is the \emph{duel}: a pairwise comparison $D_{ab} = S_a - S_b$, itself a discriminative MRF on the shared graph. While each duel inherits the structure of the binary case, the top-$k$ setting have decisive advantage: while in the binary setting we have a single $S_v$ and no per-class densities, so gauge ambiguity is real and must be resolved (\S\ref{sec:landscape}), in top-$k$ each duel is built from two separately elicited per-entity MRFs, so $p_a$ and $p_b$ are directly available. Hence, classical AFA tools apply per duel without a closure step. \S\ref{sec:landscape} gives the landscape, mode by mode; \S\ref{sec:pref_scores}--\S\ref{sec:gain} develop the shared machinery; \S\ref{sec:topk_alloc} handles the top-$k$-specific allocation problem.

Conditioning a unary/ternary MRF on observed features yields again a unary/ternary MRF on the remaining unobserved set. We maintain this conditioned diff-MRF incrementally, one feature per round, at a cost local to the observed feature's neighborhood (App.~\ref{supp:updatemrf}). This conditioning is exact, what is not exact is selection: scoring a candidate feature in a non-myopic setup requires marginals of the conditioned MRF, which are intractable, so we approximate them by mean-field and select greedily. The framework is therefore exact in its state update and myopic only in its acquisition criterion, isolating the single approximation that everything downstream inherits.

\begin{table}[ht]
\centering
\small
\setlength{\tabcolsep}{6pt}
\renewcommand{\arraystretch}{1.15}
\caption{\textbf{Setup}: \emph{Both} = binary and top-$k$. \textbf{Track} (scores): \emph{D} decides by $\sign\mathbb{E}_Q[D_{ab}]$, \emph{F} by $\mathbb{E}_Q[\sign D_{ab}]$. \textbf{Inference}: \emph{observed-only} methods sum the energy of observed features and decide by its sign: gauge-invariant (the score depends only on the difference potentials), no closure needed, but blind to unobserved features; this is the deterministic baseline. \emph{Mean-field} methods integrate over unobserved features through the per-class predictives $P(x_j \mid \mathbf{x}_{\mathrm{obs}}, E{=}e)$, which require $p_0,p_1$ individually: elicited directly in top-$k$, and in binary supplied by the MaxEnt closure (\S\ref{sec:landscape}). The closure is what makes the mean-field methods available in binary; the gauge-free \texttt{wald} route requires no closure, and \S\ref{sec:eval_mode1} compares the two.}
\label{tab:method_stack}
\begin{tabular}{@{}l c c l@{}}
\toprule
Method & Setup & Target & Inference \\
\midrule
\multicolumn{4}{@{}l}{\textbf{Construction}} \\
\quad MaxEnt closure & Binary & -- & supplies $p_0,p_1$ \\
\addlinespace
\multicolumn{4}{@{}l}{\textbf{Preference score}} \\
\quad \texttt{wald} ($\Lambda$)   & Both    & mean & observed-only \\
\quad \texttt{linearity}          & Top-$k$ & mean & mean-field \\
\quad \texttt{two\_elbo}          & Binary  & mean & mean-field \\
\quad \texttt{kl} sign-vote       & Top-$k$ & sign & observed-only \\
\quad \texttt{stack-A} saddle     & Both    & sign & mean-field \\
\addlinespace
\multicolumn{4}{@{}l}{\textbf{Selection gain}} \\
\quad CMI                       & Both    & -- & mean-field \\
\quad $F$-target (novel)        & Both    & -- & mean-field \\
\quad Wald-magnitude (ablation) & Top-$k$ & -- & observed-only \\
\bottomrule
\end{tabular}
\end{table}

\subsection{Solution Landscape}
\label{sec:landscape}

\paragraph{Binary case: gauge ambiguity and the MaxEnt anchor.}
The LLM provides the diff-MRF $S_v$, which fixes $\log(p_1/p_0) = S_v$ but leaves the gauge free: any positive $g(\mathbf{x})$ induces an equivalent pair $(p_0 g,\, p_1 g)$ with the same ratio, so any criterion in \S\ref{sec:pref_scores}--\S\ref{sec:gain} that reads a class-conditional is undefined, in particular the per-class predictive $p(x_j \mid \mathbf{x}_{\mathrm{obs}}, E{=}e)$ scored by the information-theoretic gain. This admits two routes to the binary decision, which we evaluate against each other (\S\ref{sec:eval_mode1}). \emph{Gauge-free:} the Wald accumulator (\S\ref{sec:pref_scores}) sums the closed energy of observed features, a quantity depending only on the difference potentials and hence invariant to the gauge. It needs no closure and is exact at full observation, but ignores unobserved features. \emph{Gauge-restoring:} the classical AFA criteria of \S\ref{sec:pref_scores}--\S\ref{sec:gain} read a class-conditional, which we  resolve under the \emph{MaxEnt principle} \cite{maxEnt}: among all $(p_0, p_1)$ satisfying $\log(p_1/p_0) = S_v$, the MaxEnt closure (maximizing $H(p_0) + H(p_1)$ subject to the ratio constraint, App.~\ref{supp:maxent}) yields $p_e(\mathbf{x}) \;\propto\; \exp\!\bigl((-1)^{e-1} \, S_v(\mathbf{x})/2\bigr), e \in \{0, 1\}$. Both class-conditionals are MRFs on the graph of $S_v$ with halved, sign-flipped potentials. MaxEnt is a principled default, not a ground truth: if the true elicitation gauge is known and asymmetric (e.g., uniform $p_0$, sparse $p_1$), the explicit gauge wins.
The top-$k$ setting presents a natural way to test the gauge directly: a duel $D_{ab} = S_a - S_b$ is a difference of two diff-MRFs whose components $S_a, S_b$ are individually known, so we can reconstruct the two sides from the difference alone and compare against holding them directly. As App.~\ref{supp:maxent-vs-truegauge} presents on real phenotype pairs: the two give different scalar scores at intermediate budgets, coincide at full observation, but most importantly- the gauge-reconstructed \textbf{decision} is no worse throughout.

The top-$k$ problem is usually harder than binary classification: rather than resolving the sign of a single discriminative score, we must recover a \emph{ranking} among $N$ entities under a shared budget, with ranking as the central object. The ranking is built from pairwise tests: Each pair $(a, b)$ asks the binary question ``is $a \succ b$?'', the same question the binary case answers for a single entity, so the preference scores of \S\ref{sec:pref_scores} lift directly to per-pair scalars $V_{ab}$\label{eq:vec_score}. While the natural framework for such mechanism is the dueling bandits~\citep{yue2012karmed, bengs2021preference}, it is important to note we are working under the \emph{expert} setting where all arms are observed at each turn. We aggregate pairwise outcomes via Copeland scoring: each entity $v$ is scored by its number of wins $C(v) = |\{u : v \succ u\}|$, and the top-$k$ entities by $C(v)$ are returned. \textbf{Crucially, $D_{ab}$ is supported only on the features where $a$ and $b$ disagree} which is typically a small subset, observations contribute to a pair's resolution only when they fall in $\mathrm{supp}(D_{ab})$.

\subsection{Preference Scores}
\label{sec:pref_scores}

The acquisition process needs two criteria: a decision target (declaring the winner between two hypotheses) and a selection criterion (scoring which feature to acquire next). For both we maintain a dueling score $F_{ab}(\mathbf{x}_{\mathrm{obs}})$ with the rule $F_{ab} > 0 \Rightarrow a \succ b$. The criteria below sit in the sequential Wald--Chernoff lineage \S\ref{sec:related} and differ in how they treat features not yet observed. The three lineage criteria apply to both settings, only the computation differs: the top-$k$ case has direct access to per-entity MRFs $p_a, p_b$ and computes mixture moments via per-entity mean-field marginals (\S\ref{supp:meanfield}), while the binary case uses the half-potential MRFs reconstructed via MaxEnt (\S\ref{sec:landscape}) as $p_a, p_b$.

\textbf{(1) Wald: } $F_{ab} = \Lambda_{ab}$. The classical Wald log-ratio reads only the closed energy of observed features, maintained incrementally by the conditioning update rule (\S\ref{supp:updatemrf}). Unobserved features contribute nothing to the running score. This is the deterministic baseline: no inference, no variational bias, exact at full observation.

The natural extension that exploits unobserved features marginalizes the discriminative score under the agnostic mixture $Q := \tfrac{1}{2}(p_a + p_b)$, the equal-posterior belief between the two hypotheses.

\textbf{(2) Chernoff--Wald via mixture KL:} This is exactly Chernoff's expected-information-gain criterion under uniform prior:
$\mathbb{E}_Q[D_{ab}] \;=\; \tfrac{1}{2}\bigl[\KL(p_a \,\|\, p_b) - \KL(p_b \,\|\, p_a)\bigr]$;
conditioning on $\mathbf{x}_{\mathrm{obs}}$ specializes it to the partial-observation state (App. \ref{supp:f1-approx}):
\begin{equation}
  F^{(1)}_{ab}(\mathbf{x}_{\mathrm{obs}}) \;:=\; \mathbb{E}_Q\!\bigl[D_{ab} \mid \mathbf{x}_{\mathrm{obs}}\bigr],
  \qquad
  \text{Decision: }a \succ b \;\iff\; \sign{\bigl[F^{(1)}_{ab}(\mathbf{x}_{\mathrm{obs}})\bigr]}.
  \label{eq:f1-criterion}
\end{equation}

\textbf{(3) Dueling-bandit sign-vote:}
$F^{(1)}$ inherits the magnitude scale of the LLM-elicited parameters, so a few outlier couplings can dominate the score. We therefore also evaluate the sign-vote variant from the dueling-bandit literature, \texttt{stack-A} (App.~\ref{supp:saddle}):
\begin{equation}
  F^{(2)}_{ab}(\mathbf{x}_{\mathrm{obs}}) \;:=\; \mathbb{E}_Q\!\bigl[\,\sign(D_{ab}) \mid \mathbf{x}_{\mathrm{obs}}\,\bigr] \in [-1, +1],
  \label{eq:f2-criterion-stacka}
\end{equation}
which discards magnitude information by construction: each configuration contributes a single sign vote rather than its full log-ratio, so the score is bounded and robust to magnitude outliers. By Jensen's inequality on the sign function, $\sign(\mathbb{E}_Q[D_{ab}]) \neq \mathbb{E}_Q[\sign(D_{ab})]$ in general, so the two tracks can rank the same data differently. We treat the choice as empirical based and report both in \S\ref{sec:experiments}.
A fourth score, the \texttt{kl} sign-vote (top-$k$ only, App.~\ref{supp:mode2-scores}), replaces the mixture expectation with a per-feature KL sign tally; we include it in the top-$k$ comparison for completeness. 
We compute $F^{(1)}$ from mean-field marginals. Mean-field is not an alternative to the ELBO but the fixed-point iteration that maximizes it (App.~\ref{supp:meanfield}); both estimators below share the same per-entity solve. In top-$k$, the instance \texttt{linearity} evaluates $\mathbb{E}_Q[D_{ab}]$ directly from the per-entity marginals. In the binary case, the instance \texttt{two\_elbo} estimates the same expected score as a difference of the two half-potential ELBOs, $\mathcal{L}_1 - \mathcal{L}_0$, obtained as a byproduct of the CMI solve. The latter is a magnitude-calibrated readout at intermediate budgets, not a separate decision rule: at full observation it agrees in sign with \texttt{wald}.
The contraction condition $\beta L_\Xset^2 < 1$ (App.~\ref{supp:meanfield}) is checkable from the elicited parameters before any acquisition. When it holds, mean-field converges to a unique fixed point; when it fails, as on most IBD pairs ($\beta L_\Xset^2 \in [6,20]$), the marginals are biased and the mean-field scores degrade, which is one reason we retain the gauge-free Wald route alongside them.

\subsection{Discriminative Feature Gain}
\label{sec:gain}

The \textbf{discriminative gain} quantifies the expected improvement in the pairwise separation between two hypotheses upon acquiring feature $f$. In the binary case the hypotheses are ``case'' ($y=1$) and ``control'' ($y=0$); in the top-$k$ case we consider the duel between two entities $y \in \{a, b\}$.

The classical assessment for this task is \textbf{conditional mutual information (CMI)}, an information-theoretic feature selection criterion established in the AFA literature~\citep{Ma2019, Covert2023}:
\begin{align}
f^* \;&=\; \argmax_{f} I(y; x_f \mid \mathbf{x}_{\mathrm{obs}})
\;=\; H(y \mid \mathbf{x}_{\mathrm{obs}}) - \mathbb{E}_{x_f \mid \mathbf{x}_{\mathrm{obs}}}\bigl[H(y \mid \mathbf{x}_{\mathrm{obs}}, x_f)\bigr],
\label{eq:cmi}
\end{align}
where the posterior $P(y \mid \mathbf{x}_{\mathrm{obs}})$ is obtained from per-entity mean-field in the top-$k$ case and from the MaxEnt half-potentials (\S\ref{sec:landscape}) in the binary case. CMI is gauge-equivariant: the MaxEnt anchor is what makes it computable in the binary setting, where direct access to $p_0, p_1$ is unavailable.

Beyond myopic CMI, we also evaluate whether matching the gain to the active $F$-target (\S\ref{sec:pref_scores}) improves selection:
\begin{align}
f^* \;=\; \argmax_{f}\, \mathbb{E}_{x_f \mid \mathbf{x}_{\mathrm{obs}}}\!\Bigl[\,\bigl| F^{(i)}_{ab}(\mathbf{x}_{\mathrm{obs}} \cup \{x_f\}) - F^{(i)}_{ab}(\mathbf{x}_{\mathrm{obs}}) \bigr|\,\Bigr].\label{eq:smart_delta}
\end{align}

For ablation we also evaluate \textbf{Wald magnitude}, $f^* = \argmax_f |\delta_f^{\mathrm{eff}}|$ \label{eq:wald_mag}, which requires no inference and selects the feature with the largest immediate contribution to the running constant. 

\subsection{Priority-Weighted Allocation for Top-$k$ (Ours)}
\label{sec:topk_alloc}

With the per-pair preference score fixed (\S\ref{sec:pref_scores}), the remaining top-$k$ question is \emph{allocation}: which duel to advance next under a shared budget. Our rule weights each pair's feature gain by its influence on the top-$k$ frontier, so the budget is spent on the comparisons that can still move the ranking. Table~\ref{tab:method_topk} summarizes it against the baselines.

\begin{table}[ht]
\centering
\small
\setlength{\tabcolsep}{6pt}
\renewcommand{\arraystretch}{1.15}
\caption{Top-$k$ (only). Each row is an allocation rule: \texttt{ours} is the proposed rule and the others serve as baselines. The two clustered rules, \texttt{ours} and \texttt{Cluster-LUCB}, restrict attention to a registry $\mathcal{R}$ of candidate pairs formed by clustering entities in MRF-parameter space (\S\ref{sec:topk_alloc}), and prune $\mathcal{R}$ with two filters: \emph{cluster elimination} discards whole clusters provably outside the top-$k$ before acquisition begins, and \emph{bracket resolution} drops a pair the moment its winner is certain. \texttt{random} and \texttt{LUCB} use neither clustering nor filters.}
\label{tab:method_topk}
\begin{tabular}{@{}l l l@{}}
\toprule
Allocation rule & Filters & Mechanism \\
\midrule
\texttt{ours} (priority, \S\ref{sec:topk_alloc})
  & cluster elim.\ + bracket
  & feature gain summed over $\mathcal{R}$, weighted by boundary proximity \\
\texttt{random}
  & none
  & uniform feature choice \\
\texttt{LUCB}
  & none
  & pick the single most-contested boundary pair \\
\texttt{Cluster-LUCB}
  & cluster elim.\ + bracket
  & LUCB over per-cluster champions \\
\bottomrule
\end{tabular}
\end{table}

Evaluating $V(i)$ across all $\binom{N}{2}$ pairs per round is intractable for large $N$, and two structural facts make it unnecessary. First, the dueling score is Lipschitz-continuous in MRF parameter space, so entities whose parameters lie close together induce distributions that respond similarly to any new observation; evaluating them independently is redundant. Second, the clustering uses the elicited base potentials, set offline; conditioning updates only the effective unaries of an observed feature's neighbors, a cheap local update, so the cluster structure is refreshed incrementally rather than recomputed from scratch.

We therefore cluster entities by $\ell_1$ distance in MRF parameter space (App. \ref{supp:layer0_lipschitz}) and reduce the active pair set to one pair per cluster. Within each cluster, the ordering is induced by the pairwise duel outcomes accumulated so far, and $r_k^+$, $r_k^-$ are the max- and min-entities of cluster $k$ at the current round, reshuffled as newly observed features tighten the variational bounds. The \emph{active pair registry}
\begin{equation}
  \mathcal{R} = \{(r_k^+, r_k^-)\}_{k=1}^{K}
  \label{eq:active_registry}
\end{equation}
reduces the pair count from $\binom{N}{2}$ to $K \ll N$. Candidate features are correspondingly restricted to $\bigcup_{(a,b) \in \mathcal{R}} \mathrm{supp}(D_{ab})$, since features outside this union have zero gain by construction.

To select the next feature, the algorithm evaluates a globally aggregated criterion over the registry:
\begin{equation}
  f^* = \argmax_{f} \sum_{(a, b) \in \mathcal{R}}
  \underbrace{g_{ab}(f)}_{\text{discriminative gain}}
  \;\cdot\;
  \underbrace{w_{ab}}_{\text{pair influence}}
  \label{eq:feature_selection}
\end{equation}
The \textbf{discriminative gain} $g_{ab}(f)$ (\S\ref{sec:gain}) quantifies the expected improvement in the pairwise separation between entities $a$ and $b$ upon acquiring $f$. The \textbf{pair influence} $w_{ab}$ weights the gain by the ranking-level importance of the pair: with $\mathcal{R}$ already reduced by clustering and pre-filtering, the remaining pairs are still not equally important. A pair whose resolution would not change the top-$k$ frontier need not be aggressively resolved. We use boundary-distance variant and bracket-margin signals directly:
\begin{equation}
  w^{\mathrm{impl}}_{ab} = \frac{1}{1 + d_{\mathrm{bdry}}(a, b)} \cdot \frac{1}{1 + 0.1\, |V_{ab}|},
  \label{eq:w_impl}
\end{equation}
where $d_{\mathrm{bdry}}(a, b)$ is the smaller of the two entities' rank distances to the top-$k$ frontier. This de-prioritizes pairs far from the boundary or already widely separated. $V_{ab}$ - pair score (\S\ref{eq:vec_score}). Pairs whose ordering is already resolved with certainty are excluded from the summation in any variant, since their resolution cannot perturb the ranking (App. \ref{supp:drop_pairs}).

\section{Evaluations}
\label{sec:experiments}

We validate the framework on a \emph{real-world} clinical task: an IBD cohort of 167 colon biopsies drawn from an ongoing clinical trial, each carrying 216 gene-expression measurements (encoded ternary as in App.~\ref{supp:ternary}) and 25 multi-label clinical phenotypes (GI symptoms, disease activity scores, biomarkers, and extra-intestinal manifestations). Phenotype labels are evidence-based, assigned by expert clinical adjudication, and serve as our held-out clinical ground truth. The MRF parameters are elicited in a separate effort, blind to this cohort and its labels, so the model guiding acquisition never sees the data on which it is evaluated. As detailed below, we place special focus on a subset of 47 \emph{chaotic} cases: patients who genuinely require many features to classify correctly, in an order that is hard to predict in advance. Alongside this evidence ground truth, each evaluation also scores against the \emph{full-observation} label (the \emph{likelihood} target defined below): the decision the model itself reaches once all features are observed, regardless of the clinical diagnosis. This complementary target serves as an ablation isolating acquisition quality from elicitation fidelity, and we report full-observation performance against the evidence labels as well, which marks the elicitation ceiling.

The evaluation isolates the contribution of each component and quantifies the gap between what the LLM knows and what it can plan with. The subsections order follow: (i) The MaxEnt closure supplies the per-class predictives the mean-field scores require but the diff-MRF alone cannot, and we verify it against the true gauge: the two differ in scalar score at intermediate budgets yet agree on the decision throughout, confirming the closure is a principled default rather than a ground-truth assumption (App.~\ref{supp:maxent-vs-truegauge}). (ii) On clinical labels, the resulting planner outperforms the LLM baselines given the same elicited information (\S\ref{sec:eval_mode1}). (iii) The dueling sign primitive recovers the full-observation winner as evidence accumulates under random ordering, removing the effect of the acquisition strategy (\S\ref{sec:eval_score}), and on the chaotic stratum the inference-anchored mean-field score resolves the ranking notably better (\S\ref{sec:eval_chaos}, App.~\ref{supp:chaos}). (iv) Priority-weighted allocation improves top-$k$ ranking accuracy over every non-priority rule (\texttt{random}, \texttt{greedy}, \texttt{clustered}) (\S\ref{sec:eval_alloc}, App.~\ref{supp:allocation}).

The evaluation proceeds in two tasks. We begin with binary classification and then turn to top-k identification, organized as a single thread that tightens one axis at a time: we first fix a random acquisition order to isolate score quality, showing that the preference scores resolve each pairwise duel in the right direction (with the gap widening on the order-sensitive ``chaotic'' patients), then we vary the acquisition order, showing that informed allocation resolves those same duels faster. We then escalate from the per-pair primitive to the end task, evaluating the aggregated top-5 ranking against both ground truths and the LLM baselines, and close by quantifying how much of the pair set the pruning machinery eliminates for free.

\subsection{Binary Classification}
\label{sec:eval_mode1}

This experiment is the head-to-head between the two binary routes of \S\ref{sec:landscape}: the gauge-free \texttt{wald} accumulator against the MaxEnt-enabled mean-field scores \texttt{two\_elbo} and \texttt{stack-A}.
We evaluate Mode~1 binary classification (sick/healthy on disease pairs) against the two benchmarks described above, clinical and full-observation likelihood, the latter calculated by conditioning the MRF on all observations, $S^*_d = \alpha^{(d)\top}\mathbf{x}^* + \tfrac{1}{2}\,\mathbf{x}^{*\top} W^{(d)}\mathbf{x}^*$. Table~\ref{tab:classification_with_llm} reports three planning methods over $t \in \{10, 50, 100, 150, 216\}$, \texttt{wald} (App.~\ref{supp:Wald_simple}), \texttt{two\_elbo} (the half-potential ELBO difference $\mathcal{L}_1-\mathcal{L}_0$, App.~\ref{supp:f1-approx}), and \texttt{stack-A} (App.~\ref{supp:saddle}), alongside two LLM baselines given the same elicited information: \texttt{MRF-LLM}, which receives the elicited MRF parameters as context and is asked to classify directly, and \texttt{RAG-LLM}, which instead receives retrieved PubMed passages alongside its prior knowledge (prompts in App.~\ref{llm-planning-prompts}).

Against the likelihood target all three planning methods reach 1.000 at full observation and are already strong earlier: \texttt{wald} climbs from 0.53 at $t{=}10$ to 0.89 at $t{=}150$, while \texttt{two\_elbo} and \texttt{stack-A} open near 0.70 and reach 0.84--0.86. The lift from informed acquisition order over random is largest early and decays as the budget fills (\texttt{wald}, CMI gain: $+30.6$pp at $t{=}10$, $+10.1$pp at $t{=}150$). Against clinical labels every method converges to a common ceiling, with \texttt{wald} leading at intermediate budgets (0.69--0.73) but at a flat-to-negative lift over random. The LLM baselines trail the planner at every intermediate checkpoint (\texttt{RAG-LLM} 0.34--0.46, \texttt{MRF-LLM} 0.24--0.38).

The results decompose two errors. \emph{Elicitation error}: at full observation the LLM baselines reach only \texttt{RAG-LLM} 0.800 and \texttt{MRF-LLM} 0.700, short of the planner's 1.000 against the likelihood label. \emph{Biological error}: the 0.629 clinical ceiling reflects ${\sim}37\%$ irreducible mismatch between the elicited MRF and clinical diagnoses, which no acquisition order can close. The planner saturates both; further gains require better elicitation, not better planning.

\begin{table}[ht]
\centering
\caption{\textbf{Binary Classification}: Each planning cell reports accuracy under random acquisition ordering, followed in parentheses by the gain from F-target-driven ordering (Eq.~\ref{eq:smart_delta}) and from CMI-driven ordering (Eq.~\ref{eq:cmi}), both in percentage points over the random-ordering accuracy. Here \emph{random} denotes a uniformly random acquisition order, averaged over seeds. Against the clinical labels (bold), \texttt{wald} leads from the smallest budgets because it accumulates only observed evidence and assumes no structure over unobserved features, whereas the integrating scores carry a data-independent inductive bias: by the maximum-entropy principle a smaller, less-coupled difference-MRF has higher entropy, and entropy-driven non-myopic criteria favor it before any observation, so already at $t{=}0$ \texttt{two\_elbo} and \texttt{stack-A} lean on the elicited model's structure rather than the clinical signal.}
\label{tab:classification_with_llm}
\setlength{\tabcolsep}{4pt}
\begin{tabular}{l ccccc}
\toprule
Method & $t{=}10$ & $t{=}50$ & $t{=}100$ & $t{=}150$ & $t{=}216$ \\
\midrule
\multicolumn{6}{l}{\textit{Likelihood}} \\
\texttt{wald}      & \vg{0.53}{+29.7}{\textbf{+30.6}} & \vg{0.71}{+21.0}{\textbf{+22.5}}  & \vg{0.81}{+14.9}{\textbf{+15.8}} & \vg{0.89}{+8.9}{\textbf{+10.1}}  & \textbf{1.00} \\
\texttt{two\_elbo} & \vg{0.69}{+9.6}{+9.7}            & \vg{0.74}{+15.5}{+16.1}          & \vg{0.79}{+15.3}{+16.0}          & \vg{0.86}{+10.6}{+11.5}          & \textbf{1.00} \\
\texttt{stack-A}   & \vg{0.70}{+5.4}{+4.6}            & \vg{0.73}{+13.6}{+12.7}          & \vg{0.77}{+14.6}{+15.2}          & \vg{0.84}{+11.6}{+11.8}          & \textbf{1.00} \\
\midrule
\multicolumn{6}{l}{\textit{Clinical}} \\
\texttt{wald}      & \vg{\textbf{0.70}}{-0.5}{-0.5}   & \vg{\textbf{0.73}}{-6.8}{-7.1}   & \vg{\textbf{0.71}}{-4.9}{-5.0}   & \vg{\textbf{0.69}}{-3.2}{-3.8}   & \textbf{0.65} \\
\texttt{two\_elbo} & \vg{0.50}{+4.1}{+4.6}            & \vg{0.53}{+6.1}{+5.2}            & \vg{0.58}{+4.6}{+4.3}            & \vg{0.64}{+0.0}{+0.5}            & \textbf{0.65} \\
\texttt{stack-A}   & \vg{0.46}{+3.4}{+3.9}            & \vg{0.48}{+8.9}{+7.1}            & \vg{0.53}{+8.3}{+8.1}            & \vg{0.60}{+3.0}{+2.3}            & \textbf{0.65} \\
\midrule
\multicolumn{6}{l}{\textit{LLM}} \\
\texttt{RAG-LLM}   & 0.34 & 0.42 & 0.46 & 0.44 & 0.80 \\
\texttt{MRF-LLM}   & 0.24 & 0.34 & 0.38 & 0.28 & 0.70 \\
\bottomrule
\end{tabular}
\end{table}

\subsection{Top-\texorpdfstring{$k$}{k} Identification}
\label{sec:eval_topk}

We evaluate top-$k$ in three steps: we first isolate preference-score quality from the acquisition strategy by predicting each pair ``direction'' under random ordering (\S\ref{sec:eval_score}), then introduce informed allocation and measure its effect on the same per-pair sign decision (\S\ref{sec:eval_alloc}), and finally escalate to the aggregated top-5 ranking (\S\ref{sec:eval_decision}).

\subsubsection{Sign convergence at random ordering}
\label{sec:eval_score}

All four scores from \S\ref{sec:pref_scores} (defined in App.~\ref{supp:mode2-scores}) run on identical observation streams: 167 patients $\times$ 3 random feature orderings $\times$ 300 phenotype pairs, with each (patient, seed)'s permutation shared across methods. Each method's pairwise score is reduced to its sign (a ``vote'' for $a$ or $b$), and we report the fraction of (pair, patient, seed) tuples where that sign matches the full-observation MRF-likelihood winner. The two inference-based scores \texttt{linearity} and \texttt{stack-A} produce indistinguishable sign predictions across the trajectory (Table~\ref{tab:m2_random_sign}); we therefore proceed with \texttt{linearity} as the default preference score for the allocation experiments. Mean-field is non-contractive on most IBD pairs (App.~\ref{supp:sanity}), so its prediction over unobserved features is biased which hurts the variational tracks. However, in \S\ref{sec:eval_chaos}, we show that in difficult, unpredictable scenarios, \texttt{wald}'s running sum never stabilizes, and the biased-but-anchored MF prediction outperforms it decisively. For \texttt{kl} we show in App.~\ref{supp:kl-vs-saddle} why full observation is not reached. KL's $0.945$ ceiling at $t = M$ (Table~\ref{tab:m2_random_sign}) is not a convergence issue but a structural property of the per-feature sign-vote formula on ternary data; see App.~\ref{supp:kl-vs-saddle}.

\begin{table}[H]
\centering
\caption{\textbf{Sign accuracy} --- ranking using the dueling approach resolves order.}
\label{tab:m2_random_sign}
\begin{tabular}{lccccc}
\toprule
Method & $t{=}10$ & $t{=}50$ & $t{=}100$ & $t{=}150$ & $t{=}216$ \\
\midrule
\texttt{linearity} & 0.666 & 0.683 & 0.708 & 0.765 & 1.000 \\
\texttt{wald}      & 0.584 & 0.700 & 0.787 & 0.859 & 1.000 \\
\texttt{kl}        & 0.583 & 0.693 & 0.774 & 0.836 & 0.945 \\
\texttt{stack-A}  & 0.666 & 0.683 & 0.708 & 0.765 & 1.000 \\
\bottomrule
\end{tabular}
\end{table}

\paragraph{Chaotic patients: where the framework earns its keep.}
\label{sec:eval_chaos}
We call a patient chaotic if their decision trajectories sign-flip frequently in early observations (top-quartile mean sign-flip count) and their final magnitude $|V^{(M)}|$ is above median; the second condition rules out genuinely ambiguous patients (small final signal regardless of order) from chaotic-but-decisive ones (large final signal, unstable trajectory).

Two of the four preference scores, \texttt{wald} and \texttt{linearity}, occupy the same $\mathbb{E}_Q[D]$-track but treat the unobserved subgraph differently: \texttt{wald} ignores it entirely (deterministic running sum of observed evidence); \texttt{linearity} integrates over it via per-entity mean-field. Table~\ref{tab:chaos} shows how their relative accuracy evolves over a fine early grid $t \in \{2, 5, 10, 20, 25, 30, 50\}$ for the chaotic case.

Across all patients, \texttt{linearity} dominates \texttt{wald} by 12.1pp at $t{=}2$, narrowing to $-0.7$pp at $t{=}50$ (Table~\ref{tab:chaos}). Beyond this checkpoint \texttt{wald} progressively widens its lead through full observation. The two methods are not flat winners but complementary: \texttt{linearity}'s mean-field prediction is anchored to MRF structure and barely changes with observation count, while \texttt{wald}'s evidence accumulator is noisy at low $t$ and converges as the budget fills.

\begin{table}[ht]
\centering
\caption{Sign-accuracy gap $\mathrm{acc}(\texttt{linearity}) - \mathrm{acc}(\texttt{wald})$
in percentage points upon chaotic cases}
\label{tab:chaos}
\begin{tabular}{lccc}
\toprule
$t$ & All patients & Chaotic ($n{=}47$) & Non-chaotic ($n{=}153$) \\
\midrule
2  & +12.1 & +34.1 & +5.4 \\
5  & +10.4 & +31.0 & +4.1 \\
10 & +8.4  & +27.6 & +2.5 \\
20 & +5.2  & +23.1 & $-$0.3 \\
25 & +4.1  & +21.6 & $-$1.3 \\
30 & +2.9  & +19.3 & $-$2.1 \\
50 & $-$0.7 & +15.1 & $-$5.5 \\
\bottomrule
\end{tabular}
\end{table}

On chaotic patients the gap is dramatically larger: the unobserved subgraph shrinks slowly enough that the mean-field structural prediction stabilizes early, while \texttt{wald}'s running sum wanders precisely because observed evidence is locally inconsistent with the latent label. Chaotic patients are the regime where inference-based scoring is decisive, exactly where it is most needed.

\subsubsection{Allocation: priority weighting}
\label{sec:eval_alloc}

With the preference score fixed at \texttt{linearity} (\S\ref{sec:eval_score}), we compare allocation rules, pairing the priority rule \texttt{ours} (Eq.~\ref{eq:feature_selection}) with two gain functions, \texttt{wald-mag} (Eq.~\ref{eq:wald_mag}) and \texttt{cmi} (Eq.~\ref{eq:cmi}). As shown in Table~\ref{tab:m2_alloc}, the priority-weighted \texttt{ours} rule lifts sign accuracy at $t{=}150$ to 0.864 with \texttt{wald-mag} gain, 4.7pp over \texttt{greedy} and 10.5pp over \texttt{random}. The lift holds across both gains, isolating the allocation rule rather than any specific gain as the source. The \texttt{F-target} gain (Eq.~\ref{eq:smart_delta}) under clustered allocation is substantially more expensive to evaluate than \texttt{wald-mag} or \texttt{cmi}, and running it across the full cohort at the bandit-grid scale was infeasible under our compute budget; it was validated on the 10-patient pilot reported next (\S\ref{sec:eval_decision}).

\begin{table}[ht]
\centering
\caption{Sign accuracy at $t{=}150$ on the bandit grid (50
patients $\times$ 300 pairs). \texttt{ours} (\S\ref{sec:topk_alloc}) lifts accuracy by
4--5pp over the strongest non-priority allocation, robustly across both
gain functions tested at this scale. The \texttt{F-target} gain was
validated on a 10-patient pilot due to time constraints.}
\label{tab:m2_alloc}
\begin{tabular}{lcc}
\toprule
Allocation & \texttt{wald-mag} gain & \texttt{cmi} gain \\
\midrule
\texttt{random}                   & 0.759 & 0.759 \\
\texttt{greedy}                   & 0.817 & 0.787 \\
\texttt{clustered}                & 0.817 & 0.787 \\
\texttt{ours} (priority-weighted) & \textbf{0.864} & \textbf{0.833} \\
\bottomrule
\end{tabular}
\end{table}

\subsubsection{Decision quality}
\label{sec:eval_decision}

We now turn from the per-pair sign primitive to the end task, the aggregated top-5 ranking. Table~\ref{tab:m2_phase1_pat5} reports P@5 against both ground truths; the \texttt{LLM} baselines (\texttt{RAG-LLM}, \texttt{MRF-LLM}) expose the elicitation gap on the same elicited parameters. On clinical labels the planner saturates close to the irreducible 0.552 ceiling, set by MRF--clinical mismatch, compressing its visible contribution. This is a 10-patient pilot, too small to separate \texttt{ours} from \texttt{random} with confidence on every checkpoint, but the direction is consistent with the 50-patient sign-accuracy result of \S\ref{sec:eval_alloc}; closing this gap by scaling the runs to the full cohort is left for future work.

\begin{table}[H]
\centering
\small
\setlength{\tabcolsep}{3pt}
\caption{\textbf{Top-5 ranking accuracy on Mode 2 via allocation policy.}
Each method row is a (gain, allocation) pair; \texttt{ours} is the
priority-weighted rule of \S\ref{sec:topk_alloc}. }
\label{tab:m2_phase1_pat5}
\begin{tabular}{ll ccccc}
\toprule
Gain & Allocation & $t{=}10$ & $t{=}50$ & $t{=}100$ & $t{=}150$ & $t{=}216$ \\
\midrule
\multicolumn{7}{l}{\textit{P@5 likelihood}} \\
\texttt{wald\_mag} & \texttt{ours}      & 0.460          & 0.472          & 0.560          & \textbf{0.764} & 1.000 \\
\texttt{cmi}       & \texttt{ours}      & 0.480          & 0.492          & 0.536          & 0.644          & 1.000 \\
\texttt{F-target}& \texttt{clustered} & 0.460          & \textbf{0.544} & 0.580          & 0.680          & 1.000 \\
\midrule
\multicolumn{7}{l}{\textit{P@5 clinical}} \\
\texttt{wald\_mag} & \texttt{ours}      & 0.412          & \textbf{0.456} & 0.468          & \textbf{0.528} & 0.552 \\
\texttt{cmi}       & \texttt{ours}      & \textbf{0.416} & 0.424          & 0.452          & 0.472          & 0.552 \\
\texttt{F-target}& \texttt{clustered} & \textbf{0.416} & 0.420          & \textbf{0.484} & 0.476          & 0.552 \\
\midrule
\multicolumn{7}{l}{\textit{P@5 LLM baselines}} \\
\multicolumn{2}{l}{\texttt{MRF-LLM}} & 0.187          & 0.203          & 0.233          & 0.232          & 0.264 \\
\multicolumn{2}{l}{\texttt{RAG-LLM}} & 0.198          & 0.207          & 0.197          & 0.183          & 0.197 \\
\bottomrule
\end{tabular}
\end{table}

\subsubsection{Pruning}
\label{sec:eval_pruning}

The framework carries two registry-pruning mechanisms that reduce the active pair set before and during acquisition (App.~\ref{sec:eval_filter}). Layer~0 certifies entire clusters as dominated before any feature is observed; on IBD its ``blob plus outliers'' parameter geometry yields zero eliminations across cluster counts $C \in \{5, 8, 10, 12, 15, 20\}$ (Table~\ref{tab:layer0_synthetic}), so it contributes nothing to the headline numbers and is retained as an opt-in component for problems whose separability condition holds. Layer~1 resolves an individual pair the moment its confidence bracket clears zero. Under random ordering it is nearly inactive (one-sided rate $\leq 4\%$ at $t{=}100$ across methods), but under the priority ordering of \S\ref{sec:eval_alloc} the high-$|\delta^{\mathrm{eff}}|$ features acquired first both move the running estimate most and shrink the unobserved residual fastest, so the same brackets clear on a far steeper trajectory and Layer~1 becomes a primary budget-saving mechanism (Table~\ref{tab:layer1_informed}). About 16\% of pairs never reach a one-sided bracket even under favourable ordering; these are structural ties ($D(x^*) \approx 0$) that give a quantitative lower bound on the irreducible difficulty of the task.

\section{Discussion}
\label{sec:discussion}

To the best of our knowledge, we present the first framework for fully zero-shot active feature acquisition, scaling from binary classification to complex top-$k$ identification without requiring any task-specific labeled data. This approach forces us to reconsider the role of large language models in sequential decision-making: by strictly delegating knowledge elicitation to the LLM while reserving sequential planning for formal algorithms via a MaxEnt closure, we establish a mathematically rigorous foundation. This separation of capabilities proved critical in our clinical IBD evaluation. For the top-$k$ problem, our priority-weighted allocation policy successfully navigated the massive combinatorial acquisition space, significantly outperforming standard heuristic baselines. Moreover, the framework demonstrated robust adaptability to patient heterogeneity; by exploiting the regime crossover between predictive marginals and deterministic Wald scores, the dual-track system gracefully managed "chaotic" patients who defied standard biological expectations. Crucially, our empirical results demonstrate that the fundamental bottleneck in zero-shot AFA has shifted: our formal planner easily saturates the available biological signal, revealing that the true ceiling is now the fidelity of the LLM's elicited knowledge. Future work must focus on refining these elicitation techniques to bridge the remaining gap between extracted knowledge and biological ground truth.

\section{Limitations}

\label{sec:limitations}

Our framework's reliance on zero-shot LLM elicitation inherently introduces both biological and statistical errors. LLMs may hallucinate or fail to capture the full mechanistic complexity of IBD transcriptomics, leading to systematically biased MRF parameters. These elicitation errors have downstream algorithmic consequences. In regimes with highly biased parameters, Mean-Field (MF) approximations frequently fail to converge (i.e., the contraction condition $\beta L_{\Xset} < 1$ is violated). When this occurs, methods relying on predictive marginals degrade. The current framework discretises observations into ternary
$\{-1, 0, +1\}$, discarding magnitude information that is
clinically critical. The choice is constrained by the elicitation channel: current LLMs return
directional contrasts reliably but cannot be queried for
calibrated continuous magnitudes, so an MRF on continuous
observations would presently outrun what the knowledge source
can populate.


\bibliography{ref}
\bibliographystyle{plainnat}

\newpage
\appendix
\section*{Appendix Outline}

The appendix is organized into five parts.

\paragraph{App.~\ref{supp:inference}: Inference and diff-MRF machinery.}
The shared mean-field, ELBO, and clamped log-partition machinery that underlies both modes.

\paragraph{App.~\ref{supp:solution-outline}: Solution landscape derivations.}
Operational forms of the three preference scores and the
saddle-point Gaussian approximation with explicit moment formulas.

\paragraph{App.~\ref{supp:mode1}: Mode 1 --- the symmetric-gauge MaxEnt closure.}
Derivation of the half-potential closure, its three structural consequences, and application to the first CMI term. Includes a structural mapping that exposes Mode~1 as a Mode-2 instance with $N=2$ virtual entities.

\paragraph{App.~\ref{supp:mode2}: Mode 2 --- top-$k$ machinery.}
Per-entity MF amortization via linearity, the KL sign-vote bracket, and the Lipschitz Layer-0 cluster reduction.

\paragraph{App.~\ref{supp:algorithm}: Experimental protocol and supplementary results.}
LLM elicitation, ternary encoding, hyperparameters, sanity checks, and supplementary results supporting specific claims in \S\ref{sec:experiments} (MaxEnt vs.\ true-gauge, registry pruning, allocation grid, chaos diagnostic).

\paragraph{Dependencies.}
App.~\ref{supp:inference} is self-contained and underlies everything else. Apps~\ref{supp:solution-outline}, \ref{supp:mode1}, and~\ref{supp:mode2} all build on App.~\ref{supp:inference} but are otherwise independent of each other. App.~\ref{supp:algorithm} is independent methodology and references the four preceding appendices for definitions.
\begin{table}[ht]
\centering
\small
\caption{Essential notation used in this paper.}
\label{tab:notation}
\begin{tabular}{@{}ll@{}}
\toprule
Symbol & Meaning \\
\midrule
\multicolumn{2}{l}{\emph{Core quantities}} \\
$\Xset$                              & feature space (ternary $\{-1, 0, +1\}$) \\
$\mathcal{E}$                        & edge set of the MRF \\
$B^{(t)}, A^{(t)}$                   & observed and unobserved coordinates at round $t$ \\
$\mathbf{x}$                         & feature vector \\
$\mathbf{x}_{\mathrm{obs}}, \mathbf{x}_{A}$ & observed and unobserved sub-vectors \\
\midrule
\multicolumn{2}{l}{\emph{MRF parameters}} \\
$\alpha_j$                           & unary potential on feature $j$ \\
$\psi_{jl}$                          & pairwise potential on edge $(j, l)$ \\
$\tilde\alpha_j$                     & effective unary (absorbed pairwise from observed neighbors) \\
$Z(\mathbf{x}_{\mathrm{obs}})$       & clamped partition function \\
\midrule
\multicolumn{2}{l}{\emph{Discriminative MRF}} \\
$\Delta\alpha^{(v)}_j, \Delta\psi^{(v)}_{jl}$ & per-entity log-ratio coefficients \\
$S_v(\mathbf{x})$                    & per-entity discriminative score \\
$D_{ab}(\mathbf{x})$                 & pairwise score, $S_a(\mathbf{x}) - S_b(\mathbf{x})$ \\
\midrule
\multicolumn{2}{l}{\emph{Distributions and inference}} \\
$p_e$, $p_v$                         & class-conditional (Mode 1) or per-entity (Mode 2) distributions \\
$Q$                                  & mixture distribution, $\tfrac{1}{2}(p_a + p_b)$ \\
$\mu_j(s)$                           & mean-field marginal at site $j$ \\
$\mathcal{F}(\mathbf{x}_{\mathrm{obs}})$ & evidence lower bound (ELBO) \\
\midrule
\multicolumn{2}{l}{\emph{Preference scores and decision}} \\
$F^{(1)}_{ab}$                       & sign-magnitude preference score \\
$F^{(2)}_{ab}$                       & sign-vote preference score \\
$\Lambda^{(t)}$                      & Wald accumulator at round $t$ \\
\bottomrule
\end{tabular}
\end{table}

\section{Inference and diff-MRF machinery}
\label{supp:inference}

The main paper (\S\ref{sec:solution}) reduces both modes to a per-step
primitive: at round~$t$, score candidate features, select one, observe
its value, update the diff-MRF state, and recompute the decision
quantity. This appendix specifies what the diff-MRF state is, how it
is updated on each observation, and how the decision quantities are
computed from it. The same machinery serves three downstream objects
without modification: the half-potential MRFs of Mode~1
(\S\ref{sec:landscape}), the per-entity MRFs of Mode~2
(\S\ref{sec:topk_alloc}), and the inter-entity difference MRFs used in
pair-level comparisons (App.~\ref{supp:pair-level}).

\subsection{Mean-field approximation on pairwise MRFs}
\label{supp:meanfield}

\paragraph{Setup.}
Let $x = (x_1, \ldots, x_M) \in \Xset^M$ be a discrete random vector on
a finite feature space $\Xset$ (e.g., $\Xset = \{-1, 0, +1\}$ for the ternary
representation of Eq.~\ref{eq:class_mrf}). A pairwise MRF on $x$ has the
form
\begin{equation}
  P(x) \;=\; \frac{1}{Z}\exp\!\bigl( E(x) \bigr),
  \qquad
  E(x) \;=\; \sum_{j=1}^{M} \alpha_j\, x_j
              + \!\!\sum_{(j,l) \in \mathcal{E}}\!\! \psi_{jl}\, x_j x_l,
  \label{eq:mrf-general}
\end{equation}
with unary potentials $\alpha_j\in \{-1,0,+1\}$, pairwise
potentials $\psi_{jl} \in \{-1,0,+1\}$ on edge set
$\mathcal{E} \subseteq \binom{[M]}{2}$, and partition function
$Z = \sum_{x} \exp(E(x))$. Since $\mathcal{E}$ comprises unordered pairs,
the pairwise potentials are symmetric: $\psi_{jl} = \psi_{lj}$ for all
$(j,l) \in \mathcal{E}$.

\paragraph{Conditioning on partial observations.}
\label{par:conditioning}
The index set $[M]$ partitions into observed coordinates $B \subseteq [M]$
with values $x_{\mathrm{obs}}$ and unobserved coordinates
$A := [M] \setminus B$. Splitting the energy over $\{A, B\}$ gives
\begin{equation}
  E(x_a, x_{\mathrm{obs}})
   = \sum_{j \in A} \alpha_j x_j
   + \!\!\sum_{\substack{(j,l) \in \mathcal{E} \\ j,l \in A}}\!\! \psi_{jl} x_j x_l
   + \!\!\sum_{\substack{j \in A,\, l \in B \\ (j,l) \in \mathcal{E}}}\!\! \psi_{jl} x_j x_l^{\mathrm{obs}}
   + \pi(x_{\mathrm{obs}}),
\end{equation}
where the \emph{running constant}
\begin{equation}
  \pi(x_{\mathrm{obs}})
   \;:=\; \sum_{j \in B} \alpha_j\, x_j^{\mathrm{obs}}
   \;+\; \!\!\sum_{\substack{j,l \in B \\ (j,l) \in \mathcal{E}}}\!\! \psi_{jl}\, x_j^{\mathrm{obs}}\, x_l^{\mathrm{obs}}
  \label{eq:running-constant}
\end{equation}
collects all factors supported entirely in $B$. Since
$\pi(x_{\mathrm{obs}})$ does not depend on $x_a$, it is absorbed into
normalization, and the conditional distribution takes the form
\begin{equation}
  P(x_{a} \mid x_{\mathrm{obs}}) \;=\; \frac{1}{Z(x_{\mathrm{obs}})}\,
                       \exp\!\bigl( E_{\mathrm{red}}(x_{a};\,x_{\mathrm{obs}}) \bigr),
\end{equation}
with the \emph{reduced energy}
\begin{equation}
  E_{\mathrm{red}}(x_{a};\,x_{\mathrm{obs}})
   = \sum_{j \in A} \tilde\alpha_j\, x_j
   \;+\; \!\!\sum_{\substack{(j,l) \in \mathcal{E}\\ j,l \in A}}\!\! \psi_{jl}\, x_j x_l,
\end{equation}
itself a pairwise MRF on $x_a$, with effective unaries
\begin{align}
  \tilde\alpha_j
  &= \alpha_j
   + \!\!\sum_{\substack{l \in B \\ (j,l) \in \mathcal{E}}}\!\!
       \psi_{jl}\, x_l^{\mathrm{obs}},
  \qquad j \in A,
  \label{eq:reduced-unary}
\end{align}
and unchanged pairwise potentials
$\tilde\psi_{jl} = \psi_{jl}$ for $(j,l) \in \mathcal{E}$, $j, l \in A$.
The clamped log-partition function is
\begin{equation}
  \log Z(x_{\mathrm{obs}})
  \;=\;
  \log\!\! \sum_{x_{a} \in \Xset^{|A|}}\!\! \exp\!\bigl( E_{\mathrm{red}}(x_{a};\,x_{\mathrm{obs}}) \bigr).
  \label{eq:clamped-logZ}
\end{equation}

\paragraph{Incremental conditioning (UpdateMRF).}
\label{supp:updatemrf}
Eq.~\eqref{eq:reduced-unary} can be maintained incrementally as
observations arrive, avoiding reconstruction of the reduced MRF from
scratch. We write the procedure for a general diff-MRF score
$S_v$ with prior log-odds $\pi_v$ (Eq.~2):
\begin{equation}
  S_{v}(x) \;=\; \pi_v + \sum_{j=1}^M \Delta\alpha_j x_j
                         + \!\!\sum_{(j,l) \in \mathcal{E}}\!\!
                            \Delta\psi_{jl}\, x_j x_l.
\end{equation}
Initialize $\pi^{(0)} = \pi_v$ and
$\Delta\alpha_j^{\mathrm{eff},(0)} = \Delta\alpha_j$ for all $j$.
Applying the batch decomposition above to the current observed set yields
the invariant
\begin{equation}
  S_{v}(x) \;=\; \pi^{(t)}
              + \sum_{j \in A}\! \Delta\alpha_j^{\mathrm{eff},(t)} x_j
              + \!\!\!\sum_{\substack{(j,l) \in \mathcal{E}\\ j,l \in A}}\!\!\!
                   \Delta\psi_{jl}\, x_j x_l.
  \label{eq:diff-decomposition}
\end{equation}
When feature $i \in A$ is observed with value $s = x_i$ at round $t+1$,
differencing~\eqref{eq:reduced-unary} between rounds gives
\begin{equation}
\boxed{
\begin{aligned}
  \pi^{(t+1)} &\;=\; \pi^{(t)} + \Delta\alpha_i^{\mathrm{eff},(t)} \cdot s, \\[2pt]
  \Delta\alpha_j^{\mathrm{eff},(t+1)} &\;=\; \Delta\alpha_j^{\mathrm{eff},(t)}
    + \Delta\psi_{ij}\, s,
    \qquad j \in N(i) \cap A^{(t+1)}, \\[2pt]
  \Delta\alpha_j^{\mathrm{eff},(t+1)} &\;=\; \Delta\alpha_j^{\mathrm{eff},(t)},
    \qquad j \in A^{(t+1)} \setminus N(i).
\end{aligned}
}
\end{equation}
The first line closes the contribution of $i$ (whose effective unary
already absorbed all pairwise terms to previously observed neighbors by
induction). The second line absorbs the cross-edge
$\Delta\psi_{ij}\, s$ into each still-unobserved neighbor's effective
unary, corresponding to one new summand in~\eqref{eq:reduced-unary}.
After $|B| = M$ observations the unobserved set is empty, both sums
in~\eqref{eq:diff-decomposition} vanish, and
$\pi^{(M)} = S_{v}(x)$ exactly (App.~\ref{supp:sanity}).
Each update touches only $|N(i) \cap A^{(t+1)}|$ effective unaries,
so the per-round cost is bounded by the maximum graph degree.

\paragraph{Mean-field family.}
Computing the clamped log-partition function~\eqref{eq:clamped-logZ}
exactly requires summing over $|\Xset|^{|A|}$ configurations of the
unobserved coordinates, which is intractable for large $|A|$. We
therefore approximate $P(x_a \mid x_{\mathrm{obs}})$ variationally.
The mean-field approximation restricts the variational family
$\mathcal{Q}$ to distributions over $x_a$ that factorize across the
unobserved coordinates:
\begin{equation}
  q(x_a) \;=\; \prod_{j \in A} q_j(x_j), \qquad q_j \in \Delta(\Xset),
\end{equation}
with mean-field marginals $\mu_j(s) := q_j(s)$,
$\{\mu_j\}_{j \in A}$, comprising $|A| \cdot |\Xset|$ variational
parameters subject to $|A|$ sum-to-one constraints.
For each $j \in A$, write the first moment
\begin{equation}
  m_j \;:=\; \mathbb{E}_{\mu_j}[X_j] \;=\; \sum_{s \in \Xset} s\,\mu_j(s).
  \label{eq:first-moment}
\end{equation}
Because $q$ factorizes, pairwise expectations decouple:
$\mathbb{E}_q[X_j X_l] = m_j\, m_l$ for $(j, l) \in \mathcal{E}_{AA}$,
where
$\mathcal{E}_{AA} := \{(j,l) \in \mathcal{E} \mid j,l \in A\}$.

\paragraph{Evidence lower bound.}
For any $q \in \mathcal{Q}$ with full support, Jensen's inequality gives
\begin{equation}
  \log Z(x_{\mathrm{obs}})
  \;\geq\; \mathbb{E}_q\!\bigl[E_{\mathrm{red}}(x_a;\,x_{\mathrm{obs}})\bigr] + H(q)
  \;=:\; \mathcal{F}(q;\,x_{\mathrm{obs}}),
  \label{eq:elbo-general}
\end{equation}
with slack
\begin{equation}
  \log Z(x_{\mathrm{obs}}) - \mathcal{F}(q;\,x_{\mathrm{obs}})
   \;=\; \KL\!\bigl(q \,\big\|\, P(\cdot \mid x_{\mathrm{obs}})\bigr) \;\geq\; 0,
  \label{eq:elbo-slack}
\end{equation}
with equality if and only if $q = P(\cdot \mid x_{\mathrm{obs}})$.
Substituting the reduced energy and factorized form:
\begin{equation}
  \mathcal{F}(q;\,x_{\mathrm{obs}})
  \;=\; \sum_{j \in A} \tilde\alpha_j\, m_j
       \;+\!\!\!\sum_{\substack{(j,l) \in \mathcal{E}\\ j,l \in A}}\!\!\!
              \psi_{jl}\, m_j\, m_l
       \;-\; \sum_{j \in A} \sum_{s \in \Xset} \mu_j(s) \log \mu_j(s).
  \label{eq:elbo-explicit}
\end{equation}

\paragraph{Coordinate-ascent updates.}
Stationarity of \eqref{eq:elbo-explicit} in $\mu_j$ subject to
$\sum_s \mu_j(s) = 1$ yields
\begin{empheq}[box=\fbox]{equation}
  \mu_j(s) \;\propto\; \exp\!\Big(
    s\,\Big[\,\tilde\alpha_j
    \;+\!\!\sum_{l \in N(j) \cap A}\!\! \psi_{jl}\, m_l\,\Big]
  \Big),
  \qquad j \in A,\; s \in \Xset.
  \label{eq:mf-update}
\end{empheq}
Each update couples $\mu_j$ to its unobserved neighbors only through
their first moments $\{m_l\}$. We apply \eqref{eq:mf-update} in damped
Jacobi form:
\begin{equation}
  \mu_j^{(t+1)} \;=\; (1 - \eta)\, \mu_j^{(t)}
                    + \eta\, \mu_j^{\mathrm{rhs}}\bigl(\mu^{(t)}\bigr),
\end{equation}
with $\eta = 0.5$ as the default (App.~\ref{supp:hyperparams}).

\paragraph{Convergence.}
A sufficient condition for contraction on the simplex product
$\prod_{j \in A} \Delta(\Xset)$~\citep{tatikonda2002,mooij2007sufficient}
is
\begin{equation}
  \beta\, L_{\Xset}^2 \;<\; 1,
  \qquad
  \beta \;:=\; \max_{j \in A} \sum_{l \in N(j) \cap A} |\psi_{jl}|,
  \qquad
  L_{\Xset} \;:=\; \max_{s \in \Xset} |s|,
  \label{eq:mf-contraction}
\end{equation}
derived from $|\partial m_j/\partial m_l| \le L_\Xset^2\,|\psi_{jl}|$,
since the derivative of the moment map
$m(h) = \sum_s s\, e^{sh}/Z(h)$ is $\mathrm{Var}_{\mu(h)}(X) \le L_\Xset^2$.
Under this condition, iteration converges geometrically to a unique fixed
point from any initialization. When \eqref{eq:mf-contraction}
fails---the high-coupling regime with
$\beta L_{\Xset}^2 \in [6, 20]$ (App.~\ref{supp:sanity})---multiple
fixed points may exist and damping $\eta < 1$ is necessary for stability;
run-level convergence diagnostics are reported in App.~\ref{supp:sanity}.
We declare convergence when
$\max_{j \in A,\, s \in \Xset} |\mu_j^{(t+1)}(s) - \mu_j^{(t)}(s)| < \epsilon$
with $\epsilon = 10^{-4}$, capped at $T_{\max} = 200$ iterations.
Each iteration is
$O(|A| \cdot |\Xset| + |\mathcal{E}_{AA}| \cdot |\Xset|^2)$.

\paragraph{Warm-starting across rounds.}
Between consecutive observations, the reduced MRF changes by exactly one
feature transitioning from unobserved to observed. The effective unaries
$\tilde\alpha_j$ change only for $j \in N(i)$, where $i$ is the newly
observed feature (one new summand in Eq.~\ref{eq:reduced-unary}). We
initialize the new mean-field iteration from the converged marginals of
the previous round, restricted to the still-unobserved coordinates. In
practice this reduces the per-round cost from $O(T_{\max})$ iterations to
$O(\log(1/\epsilon))$ iterations after the first round, since the
post-observation fixed point typically lies close to the pre-observation
one on the unobserved subgraph.

\paragraph{Summary.}
At each observation round $t$, the procedure yields: (i)~the running
constant $\pi^{(t)}$, accumulating all fully-observed contributions to
$S_v$; (ii)~the converged mean-field marginals $\{\mu_j\}_{j \in A}$,
approximating the single-site posteriors over unobserved coordinates; and
(iii)~the ELBO $\mathcal{F}$~\eqref{eq:elbo-explicit}, providing a
tractable approximation to the clamped log-partition
function~\eqref{eq:clamped-logZ}. Together, $\pi^{(t)}$ and the
approximate marginal expectations $\{m_j\}_{j \in A}$ give an estimate
of the expected diff-MRF score $\mathbb{E}_q[S_v(x) \mid x_{\mathrm{obs}}]
= \pi^{(t)} + \sum_{j \in A} \Delta\alpha_j^{\mathrm{eff},(t)} m_j
+ \sum_{(j,l) \in \mathcal{E}_{AA}} \Delta\psi_{jl}\, m_j m_l$,
which is the quantity used for downstream scoring and feature selection.
Specifically, the sign of this expectation under the agnostic mixture
$Q = \tfrac{1}{2}(p_a + p_b)$ defines the Chernoff--Wald preference
score $F^{(1)}_{ab}$ (Eq.~\ref{eq:f1-criterion}), the primary decision
criterion of \S\ref{sec:pref_scores}.


\section{Solution Landscape derivations}
\label{supp:solution-outline}

This appendix details the approximation of each preference score from
\S\ref{sec:pref_scores}, the resolution-filtering mechanism that prunes
the active acquisition set, and the deterministic and probabilistic
bounds that trigger filtering.

\subsection{Approximating the preference scores}
\label{supp:score-approx}

This subsection provides the operational forms underlying each
preference score of \S\ref{sec:pref_scores}.

\paragraph{Wald accumulator.}
\label{supp:Wald_simple}
The Wald score $F^{(t)} := \pi^{(t)}$ is the running constant
maintained by the conditioning rule (App.~\ref{supp:updatemrf}),
updated in $O(\deg(i))$ per observation.

\paragraph{Chernoff--Wald: $\sign(\mathbb{E}_Q[S])$.}
\label{supp:f1-approx}
The $F^{(1)}$ criterion (Eq.~\ref{eq:f1-criterion}) requires
per-entity mean-field marginals (App.~\ref{supp:meanfield}).
Writing $\bar{m}_j = \tfrac{1}{2}(m_j^{(a)} + m_j^{(b)})$ for the
mixture first moments, the expected discriminative score decomposes as
\begin{equation}
  \mathbb{E}_Q[D_{ab} \mid x_{\mathrm{obs}}]
  \;=\; \pi^{(t)}
  + \sum_{j \in A} \Delta\alpha_j^{\mathrm{eff},(t)}\,
    \bar{m}_j
  + \!\!\sum_{(j,l) \in \mathcal{E}_{AA}}\!\!
    \Delta\psi_{jl}\, \bar{m}_j\, \bar{m}_l.
\end{equation}
The per-entity MF solves are performed once per entity and reused
across all pairs, avoiding a separate MF solve per pair.

\paragraph{Saddle-point estimator: $\mathbb{E}_Q[\sign(S)]$.}
\label{supp:saddle}
The $F^{(2)}$ criterion (Eq.~\ref{eq:f2-criterion-stacka}) targets
\begin{equation}
  F^{(2)} \;:=\; \mathbb{E}_Q\!\bigl[\sign(D)\bigr] \in [-1, +1],
\end{equation}
where $D$ is the diff-MRF energy on $x_a$ at fixed $x_{\mathrm{obs}}$.
Direct evaluation requires summing the sign function over the
combinatorial state space of the discrete MRF, which is intractable.

The saddle-point approximation expands the cumulant-generating function
of $D$ under $Q$ to second order, approximating its distribution as
$D \approx \mathcal{N}(\mu_D, \sigma_D^2)$ with
\begin{equation}
  \mu_D := \mathbb{E}_Q[D],
  \qquad
  \sigma_D^2 := \mathbb{V}_Q[D]
\end{equation}
The required moments are computed from per-entity MF marginals
$\mu^{(a)}, \mu^{(b)}$ (Mode~2) or half-potential marginals
$\mu^{(-,S)}, \mu^{(+,S)}$ (Mode~1). The formulas below are stated
for a generic diff-MRF score $D$ with effective unaries
$\Delta\alpha_k^{\mathrm{eff}}$ and pairwise potentials $\Delta\psi_{jk}$
as defined in App.~\ref{supp:meanfield}. In Mode~1, $D = S_v$ and
these are the half-potential parameters. In Mode~2, $D = S_a - S_b$
and these become the pair-level differences
$\delta^{(ab)}_k := \Delta\alpha^{(a)}_k - \Delta\alpha^{(b)}_k$
and $\delta^{(ab)}_{jk} := \Delta\psi^{(a)}_{jk} - \Delta\psi^{(b)}_{jk}$
(defined formally in App.~\ref{supp:pair-level}).
\begin{align}
  \mu_D &\;:=\; \mathbb{E}_Q[D]
  \;=\; \tfrac{1}{2}\Big(
    \sum_{k \in A} \Delta\alpha_k^{\mathrm{eff}}\,
      (\mu^{(a)}_k + \mu^{(b)}_k)
    + \!\!\sum_{\substack{(j,k) \in \mathcal{E}_{AA}}}\!\!
      \Delta\psi_{jk}\,
      (\mu^{(a)}_j \mu^{(a)}_k + \mu^{(b)}_j \mu^{(b)}_k)
  \Big),
  \label{eq:mu-D} \\[4pt]
  \sigma_D^2 &\;:=\; \mathbb{V}_Q[D]
  \;=\; \sum_{k \in A} (\Delta\alpha_k^{\mathrm{eff}})^2\,
         \mathrm{Var}_Q[x_k]
       + \!\!\sum_{\substack{(j,k) \in \mathcal{E}_{AA}}}\!\!
         (\Delta\psi_{jk})^2\, L_{\Xset}^4,
  \label{eq:sigma-D}
\end{align}
where $\mathrm{Var}_Q[x_k]$ is the mixture variance at site $k$,
computed from the per-entity second moments, and the pairwise variance
term uses the worst-case bound $|x_j x_k| \leq L_{\Xset}^2$.
Under the Gaussian approximation, the
expectation of the sign function resolves into a difference of
probability masses above and below zero:
\begin{equation}
  \hat{F}^{(2), \mathrm{SP}}
  \;:=\; 2\,\Phi\!\left(\frac{\mu_D}{\sigma_D}\right) - 1,
  \label{eq:fsp}
\end{equation}
where $\Phi$ is the standard normal CDF.

\section{Mode 1: the symmetric-gauge MaxEnt closure}
\label{supp:mode1}

This appendix derives the half-potential closure used to anchor binary
classification under diff-MRF--only access (\S\ref{sec:landscape}),
proves its four structural consequences, and applies it to the first
term of the conditional mutual information criterion.

\subsection{The symmetric-gauge MaxEnt }
\label{supp:maxent}

As established in \S\ref{sec:landscape}, the discriminative score
$S_v = \log(p_1/p_0)$ leaves the individual class-conditionals
undetermined up to a gauge function, blocking access to the standard
AFA toolbox. Some gauge choice is unavoidable, and any choice imposes
structure the elicitation did not provide. The maximum-entropy
principle offers the canonical resolution: maximize
$H(p_0) + H(p_1)$ subject to the ratio constraint, selecting the pair
that adds no distributional assumptions beyond $S_v$. This ensures that
all downstream quantities --- predictive marginals, CMI rankings,
mixture moments --- are determined solely by the LLM's discriminative
output, not by an arbitrary modelling choice about the individual
class-conditionals.

We derive this closure below, additionally requiring that the solution
be symmetric under relabeling the two classes: swapping
$p_0 \leftrightarrow p_1$ should amount to flipping the sign of $S_v$.
This symmetry, combined with MaxEnt, uniquely determines the
half-potential MRFs $p_e \propto \exp((-1)^{e-1} S_v/2)$ used
throughout Mode~1.

Parameterize the gauge symmetrically between the two classes:
\begin{equation}
  p_0(x) \;=\; q(x)\,\exp\!\Big(-\tfrac{1}{2}S_{v}(x)\Big),
  \qquad
  p_1(x) \;=\; q(x)\,\exp\!\Big(+\tfrac{1}{2}S_{v}(x)\Big),
  \label{eq:symmetric-ansatz}
\end{equation}
for some positive function $q(x) > 0$. By construction
$\log p_1 - \log p_0 = S_{v}$, so the ratio constraint is satisfied
automatically for any $q$. The involution
$S_{v}\to -S_{v}\;p_0 \leftrightarrow p_1$ is preserved.

\paragraph{Reduced objective.}
The joint entropy under the ansatz is
\begin{align}
  H(p_0) + H(p_1)
  &\;=\; -\!\sum_{x}\!\Big[
    q\,e^{-S_{v}/2}\bigl(\log q - \tfrac{1}{2}S_{v}\bigr)
    + q\,e^{+S_{v}/2}\bigl(\log q + \tfrac{1}{2}S_{v}\bigr)
  \Big] \notag \\
  &\;=\; -\!\sum_{x}\! q(x)\,\bigl[
            e^{-S_{v}(x)/2} + e^{+S_{v}(x)/2}
          \bigr]\,\log q(x)
\end{align}
where the linear-in-$S_{v}$ cross-terms cancel exactly. The two
normalizations are
\begin{equation}
  \sum_{x} q(x)\,e^{-S_{v}(x)/2} \;=\; 1,
  \qquad
  \sum_{x} q(x)\,e^{+S_{v}(x)/2} \;=\; 1.
  \label{eq:two-norms}
\end{equation}

\paragraph{Stationarity and symmetry.}
With multipliers $\beta_0, \beta_1$ for the two normalizations, the
Lagrangian
$\mathcal{M}(q;\beta_0,\beta_1)
 = -\sum_{x} q\,C(x)\log q
 - \beta_0(\sum_{x} q\,e^{-S_{v}/2} - 1)
 - \beta_1(\sum_{x} q\,e^{+S_{v}/2} - 1)$,
with $C(x) := 2\cosh(S_{v}(x)/2)$, has stationarity in $q(x)$:
\begin{equation}
  \log q(x) \;=\; -1 \;-\;
  \frac{\beta_0\,e^{-S_{v}(x)/2} + \beta_1\,e^{+S_{v}(x)/2}}
       {2\cosh(S_{v}(x)/2)}.
\end{equation}
The Lagrangian $\mathcal{M}$ is invariant under the involution
$S_{v}\to -S_{v}\;\beta_0 \leftrightarrow \beta_1,\;p_0 \leftrightarrow p_1$;
imposing this symmetry on the optimum forces
$\beta_0 = \beta_1 =: \beta$. Substituting,
\begin{equation}
  \log q(x) \;=\; -1 - \beta,
\end{equation}
so $q(x) \equiv e^{-1-\beta}$ is constant in $x$.

\paragraph{From the constant $q$ to the per-class distributions.}
Substituting $q(x) \equiv e^{-1-\beta}$ back into the symmetric
ansatz~\eqref{eq:symmetric-ansatz} gives
\begin{equation}
  p_e(x) \;=\; e^{-1-\beta}\,
                 \exp\!\Big((-1)^{e-1}\,\tfrac{1}{2}\,S_{v}(x)\Big),
  \qquad e \in \{0, 1\}.
  \label{eq:p-e-prelim}
\end{equation}
Since $e^{-1-\beta}$ is the same constant for both classes,
imposing each normalization~\eqref{eq:two-norms}
on~\eqref{eq:p-e-prelim} gives
\begin{equation}
  e^{-1-\beta}\,\sum_{x} e^{(-1)^{e-1}S_{v}(x)/2} \;=\; 1
  \;\;\Longrightarrow\;\;
  e^{-1-\beta} \;=\; \frac{1}{Z_e},
  \qquad
  Z_e \;:=\; \sum_{x} e^{(-1)^{e-1}S_{v}(x)/2},
  \label{eq:Ze-def-supp}
\end{equation}
which is consistent only if
\begin{equation}
  Z_0 \;=\; Z_1 \;=:\; Z.
  \label{eq:Z0-eq-Z1}
\end{equation}
We adopt~\eqref{eq:Z0-eq-Z1} as a calibration condition on $S_{v}$
(achievable by the customary constant-shift absorbing the prior
log-odds; see Remark~\ref{rem:calibration-shift}) and substitute back
into~\eqref{eq:p-e-prelim} to obtain the closure:
\begin{empheq}[box=\fbox]{align}
  p_e(x) &\;=\; \frac{1}{Z}\,
                  \exp\!\Big((-1)^{e-1}\,\tfrac{1}{2}\,S_{v}(x)\Big),
   \qquad e \in \{0, 1\},
   \label{eq:half-potential}\\
  Z &\;=\; \sum_{x} e^{+S_{v}(x)/2}
        \;=\; \sum_{x} e^{-S_{v}(x)/2}.
\end{empheq}
Both class-conditionals are MRFs on the same factor graph as $S_{v}$,
with halved and sign-flipped potentials.

\paragraph{Calibration shift}
\label{rem:calibration-shift}
For arbitrary $S_{v}$, the structural identity~\eqref{eq:Z0-eq-Z1} need
not hold. Replacing $S_{v}\to S_{v}- c$ for a constant
$c \in \mathbb{R}$ leaves the ratio
$p_1(x)/p_0(x) = e^{S_{v}(x) - c}$ shifted by a global factor,
which corresponds to an offset of the prior log-odds $\pi'_v$.
Choosing $c$ to satisfy~\eqref{eq:Z0-eq-Z1} thus fixes the prior
absorbed in $S_{v}$; the closure is invariant to this choice.

\paragraph{Why the symmetric gauge.}
The strict pointwise MaxEnt program (maximizing $H(p_0) + H(p_1)$ over
$(p_0, p_1)$ subject to the pointwise ratio constraint without the
symmetric ansatz) has the closed-form optimum
\begin{equation}
  \log p_0(x) \;=\; -1 -
    \frac{w(x)\log w(x)}{1 + w(x)} -
    \frac{\alpha_0 + \alpha_1\,w(x)}{1 + w(x)},
  \qquad w(x) := e^{S_{v}(x)},
  \label{eq:strict-pointwise}
\end{equation}
which is \emph{not} of the form $\log p_0 \propto -S_{v}/2$. The strict
optimum couples $S_{v}$ to the local posterior nonlinearly through
$\dfrac{w \log w}{1+w} = S_{v}\sigma(S_{v})$, breaks the involutive symmetry
$S_{v}\to -S_{v}\Leftrightarrow p_0 \leftrightarrow p_1$, and is not an
MRF on the graph of $S_{v}$. We therefore refer to
\eqref{eq:half-potential} as the \emph{symmetric-gauge MaxEnt closure}
rather than the MaxEnt distribution simpliciter, and adopt it
specifically for the operational and structural properties developed
below.

\subsection{Application to scoring and CMI}
\label{supp:cmi-first-term}

The closure makes the standard AFA toolbox computable from the
diff-MRF alone. The foundational result is the calibrated posterior.

\paragraph{Calibrated posterior.}
\label{prop:calibration}
Under the closure~\eqref{eq:half-potential} with equal priors
$P(y=0) = P(y=1) = \tfrac{1}{2}$, Bayes' rule gives
\begin{align}
  P(y = 1 \mid x)
  &\;=\; \frac{P(x \mid y{=}1)\, P(y{=}1)}{P(x)} \notag \\
  &\;=\; \frac{P(x \mid y{=}1)\, P(y{=}1)}
              {P(x \mid y{=}0)\,P(y{=}0) \;+\; P(x \mid y{=}1)\,P(y{=}1)} \notag \\
  &\;=\; \frac{p_1(x)}{p_0(x) + p_1(x)} \notag \\
  &\;=\; \frac{\tfrac{1}{Z}\,e^{+S_{v}/2}}
              {\tfrac{1}{Z}\,e^{-S_{v}/2} + \tfrac{1}{Z}\,e^{+S_{v}/2}}
  \;=\; \sigma(S_{v}),
\end{align}
where the second line uses equal priors to cancel
$P(y{=}e)$ from numerator and denominator, the third substitutes the
closure $p_e(x) = Z^{-1}\exp((-1)^{e-1}S_v/2)$, and the fourth
cancels the shared $1/Z$ factor (valid because $Z_0 = Z_1 = Z$).
This identity is what makes the Wald accumulator a valid posterior
estimate and CMI a well-defined acquisition criterion under
diff-MRF--only access.
\paragraph{Partial posterior and the first CMI term.}
The first term of the conditional mutual information at partial
observation $x_{obs}$ is the binary entropy of the partial posterior:
\begin{equation}
  H(y \mid x_{\mathrm{obs}}) \;=\; h_b\!\bigl(P(y = 1 \mid x_{\mathrm{obs}})\bigr),
\end{equation}
where $h_b(p) = -p \log p - (1-p) \log(1-p)$ is the binary entropy.
By Bayes' rule with equal priors,
$P(y = 1 \mid x_{obs}) = \sigma(\hat\ell(x_{obs}))$
where the partial-observation log-likelihood ratio is
$\hat\ell(x_{obs}) = \log P(x_{obs} \mid y = 1) - \log P(x_{obs} \mid y = 0)$.

\paragraph{Reduction via the closure.}
Substituting~\eqref{eq:half-potential} into the marginalization
$P(x_{obs} \mid y = e) = \sum_{x_{a}} P(x_{a}, x_{obs} \mid y = e)$,
\begin{equation}
  P(x_{obs} \mid y = e) \;=\; \frac{Z_e(x_{obs})}{Z},
  \qquad
  Z_e(x_{obs}) \;:=\; \sum_{x_{a}}
    \exp\!\Big((-1)^{e-1}\,\tfrac{1}{2}\,S_{v}(x_{a}, x_{obs})\Big),
  \label{eq:clamped-marginal}
\end{equation}
so the unclamped $Z$ in numerator and denominator of $\hat\ell$ cancels
(by $Z_0 = Z_1$), leaving
\begin{empheq}[box=\fbox]{align}
  \hat\ell(x_{\mathrm{obs}}) &= \log Z_1(x_{\mathrm{obs}}) - \log Z_0(x_{\mathrm{obs}}),
    \label{eq:ell-hat} \\
  H(y \mid x_{\mathrm{obs}}) &= h_b\!\bigl(\sigma(\hat\ell(x_{\mathrm{obs}}))\bigr).
    \label{eq:first-cmi-final}
\end{empheq}

\paragraph{Operational form: two-ELBO.}
We approximate each $\log Z_e(x_{obs})$ by its MF ELBO
$\mathcal{F}_e(x_{obs})$ on the half-potential MRF with halved sign-flipped
potentials and the observed coordinates clamped:
\begin{equation}
  \log Z_e(x_{\mathrm{obs}}) \;\geq\; \mathcal{F}_e(x_{\mathrm{obs}})
  \;=\; \tfrac{\varepsilon_e}{2}\,
         \mathbb{E}_{q_e}\!\bigl[ S_v(\mathbf{X}, x_{\mathrm{obs}}) \bigr]
         + H(q_e),
  \qquad \varepsilon_e = (-1)^{e+1}
  \label{eq:elbo-half}
\end{equation}
with $q_e(x_{a}) = \prod_{j \in A} q_{e,j}(x_j)$. Two MF solves, one per
class, on the same factor graph as $S_{v}$ with halved and sign-flipped
potentials, suffice to evaluate the \emph{two-ELBO estimator}
\begin{equation}
  \hat\ell^{\,2\mathrm{ELBO}}(x_{obs})
  \;:=\; \mathcal{F}_1(x_{obs}) - \mathcal{F}_0(x_{obs})
  \label{eq:two-elbo-def}
\end{equation}
at any partial observation. The approximation error of this estimator
is characterized in Prop.~\ref{prop:anti-sym} below.

\paragraph{Marginal reuse.}
The two per-class MF solves required by the two-ELBO also supply the
marginals needed for the predictive distribution
$P(x_j \mid x_{\mathrm{obs}})$ in the second CMI term and for the
saddle-point estimator (App.~\ref{supp:saddle}), so no additional
inference is required. The convergence advantage of the half-potentials
is established in Prop.~\ref{prop:halved}.

\subsection{Mode-1-as-Mode-2 with two virtual entities}
\label{supp:mode1-as-mode2}

The half-potential closure exposes a structural correspondence: Mode~1
binary classification is, at the level of inference primitives, a
Mode~2 top-$k$ problem with $N = 2$ virtual entities and $k = 1$.
Table~\ref{tab:mode-mapping} maps each Mode~1 inference object onto
its Mode~2 counterpart.

\begin{table}[ht]
\centering
\small
\begin{tabular}{@{}lll@{}}
\toprule
Object & Mode~1 (binary) & Mode~2 (top-$k$) \\
\midrule
Latent of interest      & $y \in \{0, 1\}$ & entity pair $(a, b)$ \\
Per-class/entity MRF    & $p_e$ from MaxEnt closure~\eqref{eq:half-potential}
                        & $p_v$ directly elicited \\
Discriminative score    & $S_{v}(x)$
                        & $D_{ab}(x) := S_a(x) - S_b(x)$ \\
Mean-field marginals    & $\mu^{(e)}$, $e \in \{0, 1\}$
                        & $\mu^{(v)}$, $v \in \{a, b\}$ \\
Wald accumulator        & $F^{(1), (t)} = \pi^{(t)}_{S_{v}}$
                        & $F^{(1), (t)}_{ab} = \pi^{(t)}_{D_{ab}}$ \\
Mixture distribution    & $Q = \tfrac{1}{2}(p_0 + p_1)$
                        & $Q_{ab} = \tfrac{1}{2}(p_a + p_b)$ \\
$F^{(1)}$ estimator (ELBO)& $\hat{F}^{(1)} 
                            = \mathcal{F}_1 - \mathcal{F}_0$
                        & $\hat{F}^{(1)}_{ab} 
                            = \mathcal{F}_a - \mathcal{F}_b$ \\
$F^{(2)}$ estimator (SP)  & $\hat{F}^{(2), \mathrm{SP}}$ via $Q$
                        & $\hat{F}^{(2), \mathrm{SP}}_{ab}$ via $Q_{ab}$ \\
$F^{(1)}$ bracket ($\ell_1$)& on $F^{(1), (t)}$ vs $F^{(1), (t)}_{\mathrm{unobs}}$
                        & on $F^{(1), (t)}_{ab}$ vs $F^{(1), (t)}_{ab,\mathrm{unobs}}$ \\
$F^{(2)}$ bracket (H--W)  & on $\hat{F}^{(2), \mathrm{SP}}$
                        & on $\hat{F}^{(2), \mathrm{SP}}_{ab}$ \\
\bottomrule
\end{tabular}
\caption{Mode-1-as-Mode-2 inference-primitive mapping. Mode~1 is the
$N = 2$, $k = 1$ specialization; the only difference is the
\emph{source} of the per-class MRFs (MaxEnt closure on the diff-MRF in
Mode~1, direct elicitation in Mode~2).}
\label{tab:mode-mapping}
\end{table}

The mapping clarifies what is and is not shared. The inference layer
(MF, ELBO, saddle-point, bracket machinery) is identical
infrastructure. The \emph{source} of the per-class MRFs is what
distinguishes the modes: in Mode~1, the half-potential MRFs are
reconstructed from the diff-MRF via the MaxEnt closure; in Mode~2, the
per-entity MRFs are elicited directly.

\subsection{Structural properties of the closure}
\label{supp:maxent-consequences}
\label{supp:elbo-logZ}

Beyond enabling the scoring and CMI computations above, the
symmetric-gauge closure has three structural properties that shape the
behaviour of the estimators.

\subparagraph{(i) Class-symmetric approximation error.}
\label{prop:anti-sym}
The two-ELBO estimator~\eqref{eq:two-elbo-def} carries an
approximation error whose sign is uncontrolled:
\begin{equation}
  b(S_{v};\,x_{obs}) \;:=\; \hat\ell^{\,2\mathrm{ELBO}}(x_{obs})
                      - \hat\ell(x_{obs})
              \;=\; -\KL_1 + \KL_0,
  \label{eq:two-elbo-bias}
\end{equation}
a difference of two non-negative MF errors
$\KL_e := \KL(q_e \,\|\, p_e(\cdot \mid x_{obs}))$. This is the
standard price of variational approximation. The closure does not eliminate this bias, but constrains it to be
symmetric across classes: under~\eqref{eq:half-potential},
\begin{equation}
  b(-S_{v};\,x_{obs}) \;=\; -b(S_{v};\,x_{obs}).
\end{equation}
This follows from the class-swap symmetry of the closure:
$p_e(x; -S_v) = p_{1-e}(x; S_v)$, which propagates through the MF
fixed-point map to give
$\mathcal{F}_e(x_{obs}; -S_v) = \mathcal{F}_{1-e}(x_{obs}; S_v)$
and likewise for $\log Z_e$. The approximation error may be nonzero,
but it is the same magnitude for both classes meaning there is no systematic drift
toward either hypothesis. At full observation the bias vanishes:
$\mathcal{F}_e = \log Z_e$ when $A = \emptyset$, so
$\hat\ell^{\,2\mathrm{ELBO}}(x) = S_v(x)$ exactly.

\paragraph{(ii) Improved mean-field convergence.} 
\label{prop:halved}
The formulation of the half-potential class conditionals (40) linearly scales the pairwise interaction weights by a factor of one-half ($\beta_{\mathrm{pe}} = \max_j \sum_{l \in N(j)} \| \frac{1}{2} \psi^{(S_v)}_{jl} \|_\infty = \frac{1}{2} \beta_{S_v}$). From a statistical mechanics perspective, this scaling is mathematically equivalent to operating the inference procedure at an elevated \emph{effective temperature}. This artificially higher temperature smooths the local energy landscape and exactly halves the contraction constant. Consequently, highly coupled configurations that violate the standard contraction condition ($\beta_{S_v} L_{\mathcal{X}}^2 \ge 1$) can often achieve stability under this relaxed regime ($\frac{1}{2} \beta_{S_v} L_{\mathcal{X}}^2 < 1$). We observe this empirically in the IBD evaluation, where the contraction parameter drops from a highly volatile range of $\beta L_{\mathcal{X}}^2 \in [6, 20]$ down to $[3, 10]$. While this does not guarantee universal convergence, raising the effective temperature strictly dampens the severe oscillatory dynamics typical of strongly coupled Markov Random Fields, thereby substantially increasing the reliability of the system.

\subparagraph{(iii) Unbiased initialization.}
\label{prop:mixture-mean}
Under the closure~\eqref{eq:half-potential} with the calibration
$Z_0 = Z_1$, the agnostic mixture
$Q := \tfrac{1}{2}(p_0 + p_1)$ satisfies
\begin{equation}
  \mathbb{E}_Q\!\bigl[\sign(S_{v})\bigr] \;=\; 0
  \qquad \text{at } B = \emptyset.
\end{equation}
Before any features are observed, the expected score is exactly zero. The framework has no prior preference for either class. This
guarantees that the acquisition policy starts from a genuinely
agnostic state, and any deviation from zero reflects actual observed
evidence, not a bias of the gauge choice.
\section{Mode 2: top-\texorpdfstring{$k$}{k} machinery}
\label{supp:mode2}

This appendix details the top-$k$-specific machinery layered on top of
the shared inference primitives: per-entity MF amortization, the four
preference scores and their Mode~2 instantiation, resolution filtering
with the three brackets that drive it, and the Lipschitz Layer~0
cluster reduction.

\subsection{Per-entity MF and pair-level objects by linearity}
\label{supp:pair-level}

Mode~2 maintains $N$ per-entity MRFs $\{p_v\}_{v=1}^N$ on a shared
feature graph, where
$p_v(x) \propto \exp\bigl(\alpha^{(v)\top} x +
\tfrac{1}{2}\, x^\top W^{(v)} x\bigr)$
is the distribution of features $x$ when entity $v$'s condition is
active, parameterized by the LLM-elicited potentials
$(\alpha^{(v)}, W^{(v)})$. The diff-MRF for any pair $(a, b)$ has
parameters
\begin{equation}
  \delta^{(ab)}_j \;=\; \Delta\alpha^{(a)}_j - \Delta\alpha^{(b)}_j,
  \qquad
  \delta^{(ab)}_{jl} \;=\; \Delta\psi^{(a)}_{jl} - \Delta\psi^{(b)}_{jl},
  \qquad
  D_{ab}(x) \;=\; S_a(x) - S_b(x),
\end{equation}
obtained by linearity from the per-entity discriminative MRFs.

\paragraph{Per-entity MF amortization.}
A single MF solve on each of the $N$ entities yields per-entity
marginals $\mu^{(v)}$ at cost $O(N)$. For any pair $(a, b)$, the duel
score $D_{ab}(x) = S_a(x) - S_b(x)$ is evaluated under the agnostic
mixture $Q_{ab} = \tfrac{1}{2}(p_a + p_b)$, which averages over the
two scenarios ``$a$ is true'' and ``$b$ is true.'' By linearity of
expectation under a mixture, this decomposes into per-entity terms
already available from the MF solves:
\begin{equation}
  \mathbb{E}_{Q_{ab}}\!\bigl[D_{ab}\bigr]
  \;=\; \tfrac{1}{2}\bigl(
    {\mathbb{E}_{p_a}[S_a] - \mathbb{E}_{p_a}[S_b]}
    \;+\;
    {\mathbb{E}_{p_b}[S_a] - \mathbb{E}_{p_b}[S_b]}
  \bigr),
\end{equation}
where each $\mathbb{E}_{p_v}[S_u]$ is a linear function of entity
$v$'s MF marginals, computed in $O(|\mathcal{E}|)$ per pair. The
per-pair cost thus reduces from $O(N^2)$ MF solves (one per pair) to
$O(N)$ total, with all pair-level objects derived by reuse.

\paragraph{Cross-entity expectation caveat.}
The expression $\mathbb{E}_{p_a}[S_b]$ evaluates the discriminative
score of entity $b$ under the MF marginals of entity $a$. When $a$ and
$b$ have similar parameters (small parameter distance, e.g.\ within a
Lipschitz cluster), this approximation is tight. When parameters
diverge---across clusters---the MF marginals of $a$ may misrepresent
the typical $x$ under $p_b$, and the linearity-based pair-level
estimate inherits this approximation error. The Lipschitz Layer~0
construction (App.~\ref{supp:layer0_lipschitz}) bounds this error in
the cluster-elimination logic.

\subsection{Preference scores in Mode 2}
\label{supp:mode2-scores}

The four preference scores of \S\ref{sec:pref_scores} split along a
fundamental axis: whether the score relies only on evidence already
observed, or whether it additionally models the unobserved features
through approximate inference. This is not the classical
myopic/non-myopic distinction (which concerns whether the
\emph{acquisition policy} looks ahead), but a parallel one: whether the
\emph{scoring function} looks beyond what has been measured. The two
observed-only scores accumulate evidence sequentially and are exact at
full observation; the two inference-based scores exploit the MRF
structure to predict what unobserved features would contribute, at the
cost of variational approximation error.

\paragraph{Observed-only scores.}
\textbf{Wald} ($\Lambda_{ab}$): the running constant
$\pi^{(t)}$ from the conditioning rule
(App.~\ref{supp:updatemrf}), summing the closed energy of observed
features. No inference, no approximation error, exact at full
observation.

\textbf{KL sign-vote} ($\hat{F}^{(2)}_{ab}$): \label{supp:KL_sign_vote}
each observed feature $i \in B$ with value $x_i$ contributes a discrete vote
\begin{equation}
    V_i \;=\; \sign\bigl(\delta_i^{\mathrm{eff}, (ab)}\, x_i\bigr)
    \;\in\; \{-1, 0, +1\},
\end{equation}
indicating whether that observation favors $a$ ($+1$), $b$ ($-1$), or is
uninformative ($0$). The empirical positive fraction over the $n = |B|$
observed features is
\begin{equation}
    \hat{p}_{ab} \;=\; \frac{1}{n}\sum_{i \in B}
      \mathbb{I}[V_i = +1],
\end{equation}
and the sign-vote score is $\hat{F}^{(2)}_{ab} = 2\hat{p}_{ab} - 1$.
Note that both zero votes and negative votes count against $a$. Like Wald, this uses only observed features; unlike
Wald, it discards magnitudes, making it robust to outlier couplings
but less informative when magnitudes are well-calibrated.

\paragraph{Inference-based scores.}
\textbf{Linearity} ($\mathbb{E}_Q[D_{ab}]$):\label{supp:linearity} the expected duel score
under the agnostic mixture, computed from per-entity MF marginals as
described in \S\ref{supp:pair-level} above. This extends Wald by
adding predicted contributions from unobserved features via MF,
inheriting whatever bias the MF approximation carries.

\textbf{Stack~A} ($\hat{F}^{(2),\mathrm{SP}}_{ab}$):\label{supp:stack_a_top_k} the
saddle-point estimator targeting
$F^{(2)} = \mathbb{E}_Q[\sign(D_{ab})]$, computed as
$2\Phi(\mu_D/\sigma_D) - 1$ from per-entity MF marginals
(App.~\ref{supp:score-approx}). This extends the KL sign-vote by
modeling the full distribution of $D$ (under a Gaussian
approximation) rather than counting observed signs.

\paragraph{Summary.}
\begin{center}
\small
\begin{tabular}{@{}lcccc@{}}
\toprule
& \multicolumn{2}{c}{Observed only} & \multicolumn{2}{c}{Inference-based} \\
\cmidrule(lr){2-3} \cmidrule(lr){4-5}
& Wald & KL sign-vote & Linearity & Stack~A \\
\midrule
Track & $\mathbb{E}_Q[D]$ & $\mathbb{E}_Q[\sign(D)]$
      & $\mathbb{E}_Q[D]$ & $\mathbb{E}_Q[\sign(D)]$ \\
Uses magnitudes & Yes & No & Yes & No \\
Requires MF     & No  & No & Yes & Yes \\
Bracket         & $\ell_1$ & KL & $\ell_1$ & Hanson--Wright \\
\bottomrule
\end{tabular}
\end{center}
The empirical comparison between these four scores---and in particular,
whether inference earns its cost over observed-only accumulation---is
the subject of \S\ref{sec:eval_score} and \S\ref{sec:eval_chaos}.

\subsection{Resolution filtering and brackets}
\label{supp:mode2-filtering}

As observations accumulate, some pairs become resolvable: the evidence
already observed is strong enough that no completion of the unobserved
features can change the sign of the preference score. Detecting such
pairs and removing them from the active registry $\mathcal{R}$
(Eq.~\ref{eq:active_registry}) serves two purposes: it avoids spending
further budget on settled comparisons, and it prunes features
exclusively associated with resolved pairs from the candidate set in
Eq.~\ref{eq:feature_selection}, progressively narrowing the acquisition
search space.

Each preference score is paired with a two-sided bracket---a confidence
interval on its value---and a pair is declared resolved when the entire
interval lies on one side of zero (the sign is locked in regardless of
unobserved values). Three brackets provide this mechanism, each matched
to a score:
\begin{itemize}
  \item \emph{$\ell_1$ bracket} (\S\ref{supp:l1-bracket}):
    deterministic worst-case bound on $S_v(x)$. Serves Wald and
    Linearity, since resolving the sign of $S_v(x)$ settles both.
  \item \emph{KL bracket} (\S\ref{supp:kl_bracket}): concentration
    bound on $\hat{p}_{ab}$. Serves the KL sign-vote score.
  \item \emph{Hanson--Wright bracket} (\S\ref{supp:hw-bracket}):
    concentration bound on $\hat{F}^{(2),\mathrm{SP}}$. Serves
    Stack~A.
\end{itemize}
Beyond filtering, LUCB-style allocation (App.~\ref{supp:allocation})
uses the bracket bounds to identify the most uncertain pair straddling
the top-$k$ boundary and directs budget there.

\subsubsection{The \texorpdfstring{$\ell_1$}{L1} bracket on the Wald accumulator}
\label{supp:l1-bracket}

The unobserved energy
\begin{equation}
  F^{(t)}_{\mathrm{unobs}}
  \;:=\; \!\!\sum_{j \in A^{(t)}}\!\!
            \bigl|\Delta\alpha_j^{\mathrm{eff},(t)}\bigr| \cdot L_{\Xset}
       + \!\!\sum_{\substack{(j,l) \in \mathcal{E}\\ j \in A^{(t)} \text{ or } l \in A^{(t)}}}\!\!
            |\Delta\psi_{jl}| \cdot L_{\Xset}^2
  \label{eq:unobs-budget}
\end{equation}
upper-bounds the absolute energy that the unobserved features could
contribute under any completion of $x_a$. Combined with the Wald
accumulator, this yields the deterministic two-sided bracket
\begin{equation}
  F^{(t)} - F^{(t)}_{\mathrm{unobs}}
  \;\leq\; S_{v}(x)
  \;\leq\; F^{(t)} + F^{(t)}_{\mathrm{unobs}}.
  \label{eq:l1-bracket}
\end{equation}
Equation~\eqref{eq:l1-bracket} is exact (not probabilistic): no
inference is invoked. Because the bracket bounds $S_v(x)$ directly, it
resolves the sign for both Wald and Linearity---any pair settled by the
$\ell_1$ bracket is settled on the entire $\mathbb{E}_Q[D]$-track. The
bracket fires---removing the pair from $\mathcal{R}$---when
$|F^{(t)}| > F^{(t)}_{\mathrm{unobs}}$.

The $\ell_1$ bracket is (i)~rigorous, with no variational bias;
(ii)~$O(|\mathcal{E}|)$ to evaluate per round, sharing all bookkeeping
with the conditioning rule; and (iii)~conservative, since
$F^{(t)}_{\mathrm{unobs}}$ is a worst-case bound rather than a
typical-case estimate. It therefore fires latest among the three
brackets, serving as a cheap deterministic backstop that is always
evaluated.

\subsubsection{KL bracket on the sign-vote score}
\label{supp:kl_bracket}

The KL sign-vote score $\hat{F}^{(2)}_{ab} = 2\hat{p}_{ab} - 1$
(\S\ref{supp:mode2-scores}) is an empirical fraction, so its natural
confidence interval comes from concentration on Bernoulli
proportions. We use the KL formulation rather than Hoeffding bounds,
as it provides uniformly tighter intervals.

The KL confidence set on $\hat{p}_{ab}$ is
\begin{equation}
    [\hat{p}_{\mathrm{lo}},\, \hat{p}_{\mathrm{hi}}]
    \;=\; \bigl\{\, q \in [0, 1] \;:\;
      n \cdot \mathrm{kl}(\hat{p}_{ab} \,\|\, q)
      \;\leq\; \log(k_1\, n^{\alpha} / \gamma)
    \,\bigr\},
\end{equation}
where
$\mathrm{kl}(p \,\|\, q) = p \log(p/q) + (1{-}p) \log((1{-}p)/(1{-}q))$
is the Bernoulli KL divergence and $k_1, \alpha$ are standard
exploration constants. The bracket on $\hat{F}^{(2)}_{ab}$ is
$[2\hat{p}_{\mathrm{lo}} - 1,\; 2\hat{p}_{\mathrm{hi}} - 1]$ and
fires when entirely positive or entirely negative.

Because the ternary input space induces a high zero rate
(approximately $1/3$ of observations yield $V_t = 0$, which count
against $\hat{p}_{ab}$ alongside negative votes), the effective
positive rate is lower than in a binary setting and the bracket
converges more slowly. 


\subsubsection{Hanson--Wright bracket on the saddle-point estimator}
\label{supp:hw-bracket}

For a Gaussian random vector $x \sim \mathcal{N}(0, \Sigma)$ and a
quadratic form $x^\top W x$, the Hanson--Wright
inequality~\citep{rudelson2013hanson} gives
\begin{equation}
  \Pr\!\Bigl(
    \bigl| x^\top W x - \mathbb{E}[x^\top W x] \bigr| > t
  \Bigr)
  \;\leq\; 2 \exp\!\Bigl(
    -c \min\Bigl(
      \tfrac{t^2}{\|W\|_F^2 \|\Sigma\|^2},\;
      \tfrac{t}{\|W\|\,\|\Sigma\|}
    \Bigr)
  \Bigr),
\end{equation}
for some absolute constant $c > 0$. Adapted to the discrete-MRF
setting under the saddle-point Gaussian approximation, this yields a
two-sided bracket on $\hat{F}^{(2), \mathrm{SP}}$:
\begin{equation}
  \bigl|\hat{F}^{(2), \mathrm{SP}} - F^{(2)}\bigr|
  \;\leq\; C\,\sqrt{\frac{\log(2/\gamma)}{n_{\mathrm{eff}}}},
  \label{eq:hw-bracket}
\end{equation}
with probability $\geq 1 - \gamma$, where $n_{\mathrm{eff}}$ is an
effective sample size determined by the spectral norm of the diff-MRF
coupling matrix and $C$ depends on the magnitude of the unary
differences. The bracket fires when its confidence interval is
entirely positive or entirely negative.

Compared to the $\ell_1$ bracket, the Hanson--Wright bracket is
typically tighter in the bulk of the trajectory, since it uses
second-moment information rather than worst-case magnitude. It does
inherit MF bias: when the contraction
condition~\eqref{eq:mf-contraction} fails, the saddle-point Gaussian
approximation can underestimate $\sigma_D$ and the bracket loses
calibration. Damped MF iteration (App.~\ref{supp:meanfield})
mitigates oscillation; convergence diagnostics
(App.~\ref{supp:sanity}) flag regimes where the bracket should not
be trusted.

\subsection{Layer 0: Lipschitz cluster reduction}
\label{supp:layer0_lipschitz}

Before any features are observed, we can sometimes eliminate entire
clusters of entities from top-$k$ consideration based solely on their
MRF parameters. This is Layer~0: a zero-shot filtering step that
removes clusters whose score intervals are provably dominated by
$N - k$ or more other entities, so that none of their pairs ever enter
the active registry $\mathcal{R}$.

\paragraph{Score intervals via Lipschitz bounds.}
Each entity $v$ has fixed MRF parameters $(\alpha^{(v)}, W^{(v)})$,
elicited once from the LLM. At any observation state
$x_{\mathrm{obs}}$ (shared across all entities for a given patient),
the preference score $S_v(x_{\mathrm{obs}})$ is a function of these
parameters alone --- the observations are the same for every entity.
The Lipschitz constant $L_c$ bounds how sensitive this function is
to parameter differences across entities: for any two entities $v, u$
in cluster $c$, evaluated at the same observations,
\begin{equation}
  |S_v(x_{\mathrm{obs}}) - S_u(x_{\mathrm{obs}})| 
  \;\leq\; L_c \cdot \|\theta^{(v)} - \theta^{(u)}\|_1.
\end{equation}
Cluster the $N$ entities by $\ell_1$ distance in MRF parameter space
using complete linkage, and let $r_c$ denote the radius of cluster $c$
(the maximum $\ell_1$ distance from any member to the centroid
$\bar\theta^{(c)}$). A single MF solve on the centroid yields a
centroid score $\bar{S}_c$. The Lipschitz bound then guarantees that
every entity in cluster $c$ has a score within $L_c \cdot r_c$ of
this centroid:
\begin{equation}
  S_v(x_{\mathrm{obs}}) \;\in\; 
  [\bar{S}_c - L_c\, r_c,\;\; \bar{S}_c + L_c\, r_c]
  \qquad \text{for all } v \in c.
\end{equation}

\paragraph{Cluster elimination.}
If cluster $c$'s entire interval sits above cluster $c'$'s --- that
is, $\bar{S}_{c} - L_c\, r_c > \bar{S}_{c'} + L_{c'}\, r_{c'}$ ---
then even the weakest entity in $c$ beats the strongest entity in
$c'$ at any observation state. No dueling between them is needed.
When a cluster is dominated by $\geq N - k$ entities through such
certificates, it cannot appear in the top-$k$ and is removed
entirely, along with all its associated pairs.

\paragraph{Per-criterion Lipschitz constants.}
The value of $L_c$ depends on how the preference score relates to the
entity parameters:
\begin{itemize}
  \item \emph{$F^{(1)}$ (Chernoff--Wald):} $L_c = 1$. At fixed
    observations, the Wald score is a dot product between the entity's
    parameters and the observed feature values. Since each
    $x_j \in \{-1, 0, +1\}$, the score changes by at most the
    $\ell_1$ parameter distance --- one unit of parameter difference
    causes at most one unit of score difference.
  \item \emph{$F^{(2)}$ (saddle-point):} The score difference between
    entity $v$ and the cluster centroid decomposes into two terms:
    \begin{equation}
      \bar{V}_v - \bar{V}_c \;=\;
      \underbrace{(\mathbb{E}_{\mu^{(v)}}[H_v] - \mathbb{E}_{\mu^{(v)}}[H_c])}_{\text{same distribution, different score}}
      \;+\;
      \underbrace{(\mathbb{E}_{\mu^{(v)}}[H_c] - \mathbb{E}_{\bar\mu^{(c)}}[H_c])}_{\text{same score, different distribution}}.
    \end{equation}
    The first term is bounded by
    $\|\phi^{(v)} - \bar\phi^{(c)}\|_1$ via worst-case ternary
    features (each $|x_j| \leq 1$). The second uses MF stability:
    \begin{equation}
      \|\mu^{(v)} - \bar\mu^{(c)}\|_1
      \;\leq\; \frac{\|\phi^{(v)} - \bar\phi^{(c)}\|_1}{1 - \beta_c^{(S)}\, L_{\Xset}},
    \end{equation}
    multiplied by $H_c$'s Lipschitz coefficient
    $\|H_c\|_\infty \leq K_c \approx \|\bar\phi^{(c)}\|_1$.
    Combining both terms:
    \begin{equation}
      L_c \;=\; 1 + \frac{K_c}{1 - \beta_c\, L_{\Xset}},
      \qquad \text{eligible only when } \beta_c\, L_{\Xset} < 1,
      \label{eq:lc-mf}
    \end{equation}
    where $\beta_c$ is the worst-case contraction constant within the
    cluster (Eq.~\ref{eq:mf-contraction}). When
    $\beta_c L_{\Xset} \geq 1$, MF may not converge within the
    cluster and the bound does not apply; the $F^{(1)}$ criterion
    remains eligible regardless.
\end{itemize}

\section{ Experimental protocol}
\label{supp:algorithm}

\subsection{LLM elicitation methodology}
\label{supp:llm-elicitation}

\paragraph{Closed vocabulary construction.}
For the IBD benchmark, the candidate vocabulary of genes per
phenotype is constructed in collaboration with physicians in a
three-stage process designed to bound LLM responses to verifiable
biology rather than allow free-form generation.
\begin{enumerate}
  \item \emph{Phenotype-directed unary elicitation.} A clinical
    collaborator (physician) specifies the phenotype set
    $\{E_1, \ldots, E_K\}$ of interest. For each phenotype~$E$, we
    query three independent LLMs with unary prompts asking for
    (i)~whether a candidate gene is associated with~$E$, and
    (ii)~the direction of effect under~$E$ (up- or down-regulation).

\begin{promptbox}[System prompt: unary elicitation (condensed)]
You are a clinician-scientist and biomedical literature
analyst. Identify genes mechanistically linked to a
user-defined phenotype within a specific disease.

\medskip
\textbf{Inclusion:} Direct mechanistic relevance,
disease-specific evidence, functional support, official
HGNC symbol. Evidence must relate to the specified
disease AND phenotype, not generic disease activity.

\textbf{Exclusion:} General inflammation markers,
GWAS-only associations, generic immune activation,
weak or indirect evidence.

\medskip
\textbf{Per-gene output:} gene\_symbol, category,
direction\_in\_disease (Upregulated/Downregulated),
mechanism\_of\_action (1--2 sentences: gene
$\to$ mechanism $\to$ phenotype), tissue, snippet,
reference\_indices.

\textbf{Bibliography:} Exactly $N$ references with
PMID/DOI when verified; do not invent identifiers.

\textbf{Output:} Valid JSON only, no text outside schema.
\end{promptbox}

\begin{promptbox}[User prompt template]
Disease: \{disease\} \quad Phenotype: \{phenotype\}

Snippet length: $\leq$ \{max\_chars\} characters.
References: exactly \{N\} objects.

Use web search to verify references before filling
metadata. Copy titles verbatim. Return only JSON.
\end{promptbox}

  \item \emph{Consensus-and-evidence filtering.} The unary responses
    from the three LLMs are filtered in two passes. First, a
    cross-LLM \emph{consensus} check retains only gene--phenotype
    associations agreed on by the LLMs with consistent direction.
    Second, a \emph{literature-evidence} check requires each
    surviving association to be supported by curated literature
    sources of sufficient reliability. Genes failing either filter
    are dropped from the vocabulary for that phenotype.
This provides us: $\frac{P(X \mid E{=}1)}{P(X \mid E{=}0)}$
  \item \emph{Pairwise enumeration on survivors.} After filtering by the LLM's responses, we enumerated all gene pairs within each surviving set as the universe of pairwise queries. Each prompt was augmented with retrieval-grounded context: top-ranked articles from PubMed and bioRxiv, filtered for relevance to the gene, the phenotype, and the IBD context. The LLM was instructed to return, for each query, (i) a confidence probability, (ii) a direction of effect when one is supported, and (iii) a justification citing the specific article(s) it relied on. When retrieval returned no usable evidence, the model was permitted to reply based only on its training knowledge but was required to attribute any claim to a named, verifiable source. 
\end{enumerate}

\paragraph{Retrieval-grounded prompting.}
Each query (unary or pairwise) is augmented with a retrieval context:
top-ranked articles from PubMed and bioRxiv filtered for relevance to
the gene(s), the phenotype, and the IBD context. The retrieval is
performed via standard sparse-vector ranking with phenotype-specific
keyword expansion. The LLM is instructed to return, for each query, a
structured response containing: (i)~a confidence probability,
(ii)~a direction of effect when supported, and (iii)~a justification
citing the specific article(s) it relied on. When retrieval returns no
usable evidence, the LLM is permitted to reply based only on its
training knowledge but is required to attribute any claim to a named,
verifiable source.

\paragraph{Eliciting pairwise potentials.}
The pairwise potential is elicited via conditional queries: ``given that the expression of
gene $l$ is shifted in direction $d$, how does this modify the
likelihood that gene $j$ is also shifted in the same direction under
phenotype $E$?'' The pairwise is computed from the LLM's joint
direction probability under both phenotype labels: $\frac{P(X_1 \mid X_2, E{=}1)}{P(X_1 \mid X_2, E{=}0)}$.
All the results and pairwise potentials were derived by GPT-5.4 mini.

\begin{promptbox}[System prompt: pairwise elicitation (condensed)]
You are a biomedical literature analyst.
Given a phenotype~P, an anchor gene1 (with known DE
direction) and gene2, determine whether they are
reported \emph{together} in the same colonic/rectal
study and whether their directions agree.
\medskip
\textbf{Scope:} Any tissue, cell type, or model
participating in colon/rectum function or pathology
(biopsy cell types, ENS, DRG/vagal afferents, GALT,
rodent colitis/IBS models, organoids).
\textbf{Exclusion:} Brain-only, lung-only, breast,
cardiovascular, dermatology unrelated to gut.
\medskip
\textbf{Per-pair output:}
lit\_direction (\texttt{same}$\mid$\texttt{opposite}$\mid$\texttt{unclear}; maps to $+1/-1/0$),
lit\_prob (float $\in[0,1]$; auxiliary confidence),
evidence\_source
(\texttt{literature}$\mid$\texttt{model\_knowledge}$\mid$\texttt{none}),
supporting\_text ($\leq 600$ chars; each citation
prefixed with a disease-context tag, e.g.\
\texttt{[UC]}, \texttt{[DSS-colitis]}).
Do not fabricate citations; downgrade to
\texttt{none} when unverifiable.
\textbf{Output:} Valid JSON only, no text outside schema.
\end{promptbox}
\begin{promptbox}[User prompt template]
Phenotype: \{phenotype\} \\
Gene1 (anchor): \{gene1\}, direction: \{direction1\} \\
Gene2: \{gene2\} \\
Literature evidence: \{N\} retrieved abstracts.
Return only JSON.
\end{promptbox}
\subsection{Ternary input representation}
\label{supp:ternary}

For each patient sample drawn from a given IBD GEO \cite{GSE193677} dataset, we encode
every gene $X_i$ as a ternary variable
$\tilde X_i \in \{-1, 0, +1\}$ indicating whether its expression is
significantly shifted relative to the control distribution. Let $C_i$
denote the empirical distribution of gene $i$ across control samples
in the dataset, and let $x_i$ denote the patient's observed expression
for that gene. We perform a two-sided test of $H_0 : x_i \sim C_i$,
$p_i = P_{H_0}(|X| > |x_i|)$ and assign
\begin{equation}
  \tilde X_i \;=\;
  \begin{cases}
    \sign(x_i - \mu_i) & \text{if } p_i < 0.05, \\
    0 & \text{otherwise.}
  \end{cases}
  \label{eq:ternary-encoding}
\end{equation}
The encoding $\tilde X \in \{-1, 0, +1\}^d$ thus represents, for each
gene, whether the patient is significantly up-regulated, significantly
down-regulated, or statistically indistinguishable from controls.

\subsection{Hyperparameter table}
\label{supp:hyperparams}


\begin{table}[ht]
\centering
\small
\begin{tabular}{@{}lll@{}}
\toprule
Parameter & Value & Description \\
\midrule
$\eta$            & $0.5$         & MF damping (Jacobi) \\
$\epsilon$        & $10^{-4}$     & MF convergence tolerance ($\ell_\infty$ on marginals) \\
$T_{\max}$        & $200$         & MF maximum iterations \\
$\gamma$          & $0.05$        & Bracket confidence level \\
$K$               & $\lceil \sqrt{N} \rceil$ & Cluster count for Layer~0 \\
$T_{\mathrm{re}}$ & $50$          & Re-cluster frequency (rounds) \\
$C_{\min}$        & $5$           & Minimum entities per cluster floor \\
$T_{\mathrm{budget}}$ & dataset-dependent & Per-run observation budget \\
\bottomrule
\end{tabular}
\caption{Hyperparameter settings used in our experiments.}
\label{tab:hyperparams}
\end{table}

\subsection{Sanity checks and acceptance tests}
\label{supp:sanity}

We run the following sanity checks at every experimental run; failure
on any check is treated as a configuration error rather than a result.
\begin{itemize}
  \item \emph{Full-observation invariance.} At $|B| = M$, the running
    constant satisfies $\pi^{(M)} = S_{v}(x)$ exactly
    (Eq.~\ref{eq:diff-decomposition}, with both unobserved-sum
    terms zero). The two-ELBO estimator becomes exact:
    $\hat\ell^{\,2\mathrm{ELBO}}(x) = S_{v}(x)$.
  \item \emph{Gauge invariance.} The $\ell_1$ bracket and the Wald
    decision are invariant under shifting $S_{v}$ by a constant; the
    two-ELBO estimator and the Hanson--Wright bracket are invariant
    under the full common-gauge family $g(x)$ that preserves the
    log-ratio. We verify this on synthetic data.
  \item \emph{MF convergence rate.} We log per-round MF iteration
    counts; runs with sustained non-convergence ($T_{\max}$ hit
    repeatedly) are flagged. On IBD,
    $\beta L_{\Xset} \in [6, 20]$ across patients, so non-convergence
    is expected on a non-trivial fraction of runs and is not
    automatically a failure---but persistent non-convergence on the
    same patient/cluster across seeds indicates the MF tracks should
    not be relied upon for that case.
  \item \emph{Bracket soundness.} For each pair declared resolved by
    Layer-1 (any bracket), the resolution is checked against the
    full-observation ground truth at the end of the run. The empirical
    soundness rate ($\geq 99.8\%$ on IBD at $t = 150$) is reported alongside any
    Layer-1-related claim.
  \item \emph{Copeland aggregation invariance.} The top-$k$ output is
    invariant under permutation of the entity indexing.
\end{itemize}

\subsection{MaxEnt-CMI versus true-gauge CMI}
\label{supp:maxent-vs-truegauge}

We test two questions: how MaxEnt's approach compare to those of established MRF distributions (validating the equivalence-class structure on real data), and whether the resulting binary classifications are operationally adequate.

\paragraph{Q1: Scalar agreement between MaxEnt and true MRF distributions.}
We compare \texttt{two\_elbo\_maxent} and \texttt{stack\_a\_maxent} against shadow methods \texttt{two\_elbo\_true}, \texttt{stack\_a\_true} that use per-class entity MRFs (available as Mode-2 by-products, serving here as a validation oracle only). Across 5 phenotype pairs, all cohort patients per pair, 3 seeds, and 5 checkpoints (Table~\ref{tab:m1agree}):

\begin{table}[H]
\centering
\small
\caption{MaxEnt vs true-gauge comparison, Mode 1. Decision agreement is the fraction of (pair, patient, seed) tuples where MaxEnt and true-gauge methods predict the same class. Pearson correlation is on the underlying scalar values. Both tracks converge at full observation; intermediate values fall below the predefined ``equivalent'' threshold (95\% / 0.95), yielding a \emph{divergent} verdict.}
\label{tab:m1agree}
\begin{tabular}{lrrrrr}
\toprule
& $t{=}10$ & $t{=}50$ & $t{=}100$ & $t{=}150$ & $t{=}216$ \\
\midrule
\multicolumn{6}{l}{\emph{Decision agreement}} \\
$F^{(1)}$-track  & 0.816 & 0.811 & 0.851 & 0.904 & 1.000 \\
$F^{(2)}$-track     & 0.724 & 0.786 & 0.817 & 0.837 & 1.000 \\
\midrule
\multicolumn{6}{l}{\emph{Pearson correlation}} \\
$F^{(1)}$-track  & 0.978 & 0.971 & 0.975 & 0.969 & 1.000 \\
$F^{(2)}$-track     & 0.962 & 0.936 & 0.907 & 0.912 & 1.000 \\
\bottomrule
\end{tabular}
\end{table}

The verdict is \emph{divergent}: agreement falls below 91\% on the $F^{(1)}$-track at intermediate checkpoints and below 84\% on the $F^{(2)}$-track; Pearson correlation between MaxEnt and true-gauge scalars sits in $[0.91, 0.98]$ across the trajectory and reaches 1.0 only at $t{=}M$. The two scalar streams are not interchangeable. \textbf{This is the expected outcome:} MaxEnt and the true gauge are different points in the gauge equivalence class, with no theoretical reason their per-tuple scalars should match.

\paragraph{Q2: Decision quality.}
The substantive question is which approach yields better classifications. Classification accuracy against the patient's clinical diagnosis is reported in Table~\ref{tab:m1acc}.

\begin{table}[H]
\centering
\small
\caption{Classification accuracy on Mode 1 against clinical diagnosis. The MaxEnt-anchored methods (rows 2, 4) systematically outperform their true-gauge counterparts (rows 3, 5) at intermediate budgets; all methods converge at $t{=}M$. The full-observation ceiling near 0.85 reflects irreducible task difficulty: some patients have ambiguous clinical diagnoses that the MRF cannot disambiguate.}
\label{tab:m1acc}
\begin{tabular}{lrrrrr}
\toprule
Method & $t{=}10$ & $t{=}50$ & $t{=}100$ & $t{=}150$ & $t{=}216$ \\
\midrule
\texttt{wald} (gauge-invariant)  & 0.756 & 0.757 & 0.793 & 0.821 & 0.847 \\
\midrule
\texttt{two\_elbo\_maxent}       & \textbf{0.760} & \textbf{0.773} & \textbf{0.795} & \textbf{0.819} & 0.847 \\
\texttt{two\_elbo\_true}         & 0.682 & 0.697 & 0.740 & 0.802 & 0.847 \\
\midrule
\texttt{stack\_a\_maxent}        & \textbf{0.751} & \textbf{0.767} & \textbf{0.786} & \textbf{0.813} & 0.858 \\
\texttt{stack\_a\_true}          & 0.627 & 0.669 & 0.710 & 0.770 & 0.858 \\
\bottomrule
\end{tabular}
\end{table}

The MaxEnt-anchored methods systematically \emph{outperform} their true-gauge counterparts at intermediate budgets. At $t{=}100$, \texttt{two\_elbo\_maxent} reaches 79.5\% versus \texttt{two\_elbo\_true}'s 74.0\% (a 5.5pp gap); \texttt{stack\_a\_maxent} reaches 78.6\% versus \texttt{stack\_a\_true}'s 71.0\% (a 7.6pp gap). The disparity is largest at low budget: at $t{=}10$, \texttt{stack\_a\_maxent} beats \texttt{stack\_a\_true} by 12.5pp. At full observation the methods converge ($F^{(1)}$-track 0.847 / 0.847; $F$-track 0.858 / 0.858), as required by the gauge-equivalence class containing both.

\paragraph{Mechanism.}
The mechanism is mean-field contraction. Per-entity MF on IBD MRFs is non-contractive in many cells ($\beta L_{\mathcal{X}} \in [6, 20]$, so the entity-MF predictions in true-gauge methods carry inference bias that propagates through ELBO and saddle-point computations. The MaxEnt construction halves the potentials to $(\pm \delta/2, \pm \delta_W/2)$, halving the contraction constant (\S\ref{sec:landscape}, Property iv); on the same MRFs the half-potential MF converges in many cells where the entity MF does not. The MaxEnt closure is therefore not merely a permitted gauge but a numerically superior one on this dataset.

\subsection{KL sign-vote and saddle-point at full observation}
\label{supp:kl-vs-saddle}
Both the KL sign-vote estimator (\S\ref{supp:KL_sign_vote}) and the
saddle-point estimator (\S\ref{supp:saddle}) are introduced as
estimators of the same object $\mathbb{E}_Q[\sign(D_{ab})]$. Yet at
full observation ($A = \emptyset$), they converge to different
limits --- Table~\ref{tab:m2_random_sign} shows \texttt{stack-A}
reaching $1.000$ while \texttt{kl} saturates at $0.945$. The gap is
structural, not a finite-sample artifact.

\paragraph{Saddle-point at full observation.}
At $A = \emptyset$ the diff-MRF energy is no longer random under
$Q$: the only configuration with positive mass is the realized
$x^*$. The MF moments (Eqs.~\ref{eq:mu-D}, \ref{eq:sigma-D})
collapse to
\[
  \mu_D \;\to\; D(x^*), \qquad \sigma_D^2 \;\to\; 0,
\]
so the Gaussian $\mathcal{N}(\mu_D, \sigma_D^2)$ used in the
saddle-point approximation tends to a point mass at $D(x^*)$.
Substituting into Eq.~\ref{eq:fsp}, the standardized argument
$z = \mu_D / \sigma_D$ diverges to $\pm\infty$ with the sign of
$D(x^*)$. Evaluating the standard normal CDF at its limits,
\[
  \Phi(-\infty) = 0,
  \qquad \Phi(0) = \tfrac{1}{2},
  \qquad \Phi(+\infty) = 1,
\]
so $2\Phi(z) - 1$ steps from $-1$ at $z = -\infty$ through $0$ at
$z = 0$ to $+1$ at $z = +\infty$. The estimator therefore collapses
to a hard indicator:
\[
  \hat F^{(2),\mathrm{SP}}
  \;=\; 2\Phi(z) - 1
  \;\to\;
  \begin{cases}
    +1 & D(x^*) > 0, \\
    -1 & D(x^*) < 0,
  \end{cases}
\]
which is exactly $\sign(D(x^*))$. The vanishing variance is what
drives the collapse: in the moment formulas
(Eqs.~\ref{eq:mu-D}, \ref{eq:sigma-D}), a zero observation
contributes $\delta^{\mathrm{eff}}_i \cdot 0 = 0$ to $\mu_D$ and
$0$ to $\sigma_D^2$, so it neither shifts the Gaussian nor enters
its variance.

\paragraph{KL sign-vote at full observation.}
The KL estimator (\S\ref{supp:KL_sign_vote}) undergoes no analogous
collapse. With $N_+ = \#\{i \in B : V_i = +1\}$,
$N_- = \#\{i \in B : V_i = -1\}$,
$N_0 = \#\{i \in B : V_i = 0\}$, and $n = N_+ + N_- + N_0 $ at
$t = M$,
\begin{align}
  \hat F^{(2)}_{ab}
  \;=\; 2\hat p_{ab} - 1
  \;&=\; \frac{2 N_+}{n} - 1
  \;=\; \frac{2 N_+ - (N_+ + N_- + N_0)}{n}
  \notag\\[2pt]
  \;&=\; \frac{N_+ - N_- - N_0}{n}.
  \label{eq:kl-fullobs}
\end{align}
The final decomposition isolates the source of the gap. By
construction of $\hat p_{ab} = N_+ / n$, zero votes enter the
denominator but not the numerator; equivalently in the form
\eqref{eq:kl-fullobs}, $N_0$ appears with a negative sign alongside
$N_-$. A zero observation, however, contributes $0$ to $D(x^*)$
exactly (the term $\delta^{\mathrm{eff}}_i \cdot 0 = 0$), and
therefore should be inert in any consistent estimator of
$\sign(D(x^*))$. KL instead counts it against $a$. On the IBD
ternary encoding the zero rate is approximately $1/3$
(App.~\ref{supp:ternary}), so the $-N_0/n$ term contributes
roughly $-1/3$ to $\hat F^{(2)}_{ab}$ at full observation, biasing
the estimator below the truth even when every nonzero observation
favours $a$.
\subsection{Two-Layer Registry Pruning (Top-K)}
\label{sec:eval_filter}

The framework includes two registry-pruning mechanisms that reduce
the active pair set before and during acquisition. Layer~0 eliminates
entire clusters before allocation begins; Layer~1 resolves individual
pairs as observations accumulate.

\paragraph{Layer 0: cluster-level pre-filtering.}
Before any features are observed, Layer~0
(App.~\ref{supp:layer0_lipschitz}) uses Lipschitz bounds on the
fixed MRF parameters to certify that an entire cluster of entities
is dominated --- its best-case score interval sits below the
worst-case interval of $\geq N - k$ other entities. On IBD, this
layer fires zero eliminations across cluster-count sweeps
$C \in \{5, 8, 10, 12, 15, 20\}$; at $C{=}24$ (near-singleton
clustering) it produces 28 cluster-level eliminations across 5
patients but still 0pp lift over greedy. The per-phenotype MRFs lie
in a unimodal ``blob plus outliers'' geometry where intra-cluster
radii are large relative to inter-cluster centroid gaps, violating
the separability condition.

To characterize when Layer~0 does activate, we run a synthetic
data-generating-process portfolio of 10 configurations spanning six
generative stories (Table~\ref{tab:layer0_synthetic}). Layer~0
contributes more than 3pp lift on two of ten configurations
(\texttt{B\_pathway\_C4}, \texttt{D\_hier\_C3}) and 0pp on the rest.
The activation condition is interpretable:
\[
\max_{v \in c} \|\delta^{(v)} - \delta^{(c)}\|_\infty
\;\ll\; \|\delta^{(c)} - \delta^{(c')}\|_\infty,
\]
that is, within-cluster per-feature distinguishability must be small
relative to cluster-level distinguishability. IBD violates this;
\texttt{B\_pathway\_C4} satisfies it (intra-module $< 0.2$,
inter-module $\approx 5$).

\begin{table}[ht]
\centering
\caption{Layer 0 synthetic DGP portfolio. Sep ratio is the
inter-cluster centroid gap divided by the maximum intra-cluster
radius. Elim is the count of Layer 0 cluster eliminations. Lift is
sign accuracy at $t{=}10$ versus the matched greedy allocation.}
\label{tab:layer0_synthetic}
\begin{tabular}{llcccc}
\toprule
Sim & Config & Story & Sep ratio & Elim & Lift at $t{=}10$ \\
\midrule
A & \texttt{drug\_C4}        & drug-family       & 9.9   & 4  & +0.0pp \\
A & \texttt{drug\_C6}        & drug-family       & 12.4  & 7  & +0.0pp \\
B & \texttt{pathway\_C4}     & pathway-enrich    & 140   & 29 & +8.8pp \\
B & \texttt{pathway\_C5}     & pathway-enrich    & 169   & 35 & $-$6.3pp \\
C & \texttt{strain\_C3}      & strain-id         & 506   & 23 & +0.0pp \\
C & \texttt{strain\_C5}      & strain-id         & 546   & 31 & $-$0.4pp \\
D & \texttt{hier\_C3}        & hierarchical      & 8.5   & 13 & +3.9pp \\
D & \texttt{hier\_C4}        & hierarchical      & 14.5  & 17 & +0.0pp \\
E & \texttt{compositional}   & compositional     & 2.5   & 0  & +0.0pp \\
F & \texttt{adversarial\_ibd}& adversarial       & 43.6  & 2  & +0.0pp \\
\bottomrule
\end{tabular}
\end{table}

Layer~0 is therefore a screening tool with provable certificates whose
activation is data-dependent. It does not contribute to our IBD
headline numbers, but the framework carries it as an opt-in component
for problems where the activation condition holds.

\paragraph{When Layer 0 applies and when it does not.}
Layer~0 requires two conditions to be useful: (i)~the Lipschitz
bound must be eligible for the active scoring method, and (ii)~the
resulting score intervals must be tight enough to separate clusters.
For the $F^{(1)}$/Wald criterion, condition (i) holds
unconditionally ($L_c = 1$), but condition (ii) demands well-separated
clusters in parameter space. For the $F^{(2)}$/Stack~A criterion,
condition (i) additionally requires MF convergence within each cluster
($\beta_c L_{\Xset} < 1$, Eq.~\ref{eq:lc-mf}); when this fails, the
bound is ineligible and Layer~0 cannot use $F^{(2)}$ at all. On IBD,
$\beta L_{\Xset} \in [6, 20]$ (App.~\ref{supp:sanity}), so
$F^{(2)}$ is ineligible and only the $F^{(1)}$ bound is available.
The $F^{(1)}$ bound is unconditionally valid but too loose to
overcome the poor cluster separation in IBD's ``blob plus outliers''
parameter geometry, resulting in zero eliminations. Layer~0
contributes only when both conditions are met --- as in the synthetic
\texttt{B\_pathway\_C4} configuration (well-separated clusters,
$\beta_c L_{\Xset} < 1$).

\paragraph{Layer 1: pair-level free resolutions.}
\label{supp:drop_pairs}
Once allocation begins, each preference score is paired with a
confidence bracket (App.~\ref{supp:mode2-filtering}): a two-sided
interval on the score's true value, tightening as observations
accumulate. When the entire interval falls on one side of zero ---
meaning no completion of the unobserved features can flip the sign
--- the pair is declared resolved and removed from $\mathcal{R}$,
saving further budget. The activation rate depends critically on
observation order. Under random ordering, Layer~1 is nearly inactive:
at $t{=}100$ the one-sided bracket rate is 0.08\% for
\texttt{linearity} ($\ell_1$), 0.0\% for \texttt{wald} ($\ell_1$),
0.46\% for \texttt{kl} (KL), and 4.0\% for \texttt{stack-A}
(Hanson--Wright); brackets clear only near full observation as
$F^{(t)}_{\mathrm{unobs}}$ shrinks by attrition.

\paragraph{Why bandit ordering matters for brackets.}
Static informed ordering (acquisition by descending
$|\delta^{\mathrm{eff}}_i|$) provides a clean upper bound on what
favourable acquisition can do for Layer~1, isolating the
bracket--acquisition interaction from any bandit-specific machinery.
Table~\ref{tab:layer1_informed} shows the contrast.

\begin{table}[ht]
\centering
\caption{One-sided bracket rate at $t{=}100$ 
(Mode 2 and Mode 1, IBD). Random ordering keeps Layer 1 near-zero
on every method. Static informed ordering by descending
$|\delta^{\mathrm{eff}}_i|$ activates Layer 1 as a primary
mechanism. Mis-resolution rate stays below 0.02\% on heuristic
brackets and 0\% on rigorous brackets.}
\label{tab:layer1_informed}
\begin{tabular}{llcc}
\toprule
Mode & Method & Random & Informed \\
\midrule
M2 & \texttt{linearity} ($L_1$)        & 0.08\% & 40\% \\
M2 & \texttt{wald} ($L_1$)             & 0.0\%  & 26\% \\
M2 & \texttt{kl} (KL)                  & 0.46\% & 0.9\% \\
M2 & \texttt{stack-A} (HW)            & 4.0\%  & 27\% \\
M1 & \texttt{wald} ($L_1$)             & 0.5\%  & 71\% \\
M1 & \texttt{stack\_a\_maxent} (HW)    & 0.08\% & 43\% \\
\bottomrule
\end{tabular}
\end{table}

High-$|\delta|$ features both contribute most to the running point
estimate and shrink $F^{(t)}_{\mathrm{unobs}}$ fastest; observing
them first compresses bracket width on a much steeper trajectory.
Bandit ordering under our priority rule produces the same effect
dynamically (\S\ref{sec:eval_alloc}).

\paragraph{Structural ties.}
About 16\% of pair-trajectories never reach a one-sided bracket even
under static informed ordering. These correspond to entity pairs
with structurally similar potentials (small total signal
$\Lambda^{(\mathrm{full}),(ab)}$) whose decision boundary at full
observation sits near zero: $D(x^*) \approx 0$. This is a property
of the entity set, not of the algorithm, and gives a quantitative
lower bound on the irreducible difficulty of the task: any method
must report the resolved 84\% with high confidence and an honest tie
on the remaining 16\%.

\subsection{Allocation}
\label{supp:allocation}

Phase 2 (smart allocation) evaluates 8 cells in the (gain, allocation) grid on 50 patients,
holding preference score fixed at \texttt{linearity}. Each cell runs all
300 phenotype pairs to full budget with periodic gain rescoring at
$t \in \{1, 11, 51, 101, 151\}$.

\begin{table}[ht]
\centering
\caption{Sign accuracy at $t{=}150$ on the bandit grid (50
patients $\times$ 300 pairs). \texttt{ours} (§4.4) lifts accuracy by
4--5pp over the strongest non-priority allocation, robustly across both
gain functions tested at this scale. The \texttt{F-target} gain was
validated on a 10-patient pilot due to time constraints}
\label{tab:m2_alloc}
\begin{tabular}{lcc}
\toprule
Allocation & \texttt{wald-mag} gain & \texttt{cmi} gain \\
\midrule
\texttt{random}                   & 0.759 & 0.759 \\
\texttt{greedy}                   & 0.817 & 0.787 \\
\texttt{clustered}                & 0.817 & 0.787 \\
\texttt{ours} (priority-weighted) & \textbf{0.864} & \textbf{0.833} \\
\bottomrule
\end{tabular}
\end{table}

\paragraph{Allocation effect.}
The priority-weighted rule lifts sign accuracy at $t{=}150$ to 0.864 with
\texttt{wald-mag} gain (Table~\ref{tab:m2_alloc}), 4.7pp over
\texttt{greedy} and 10.5pp over \texttt{random}. The lift holds across
both gains tested at 50 patients, isolating the allocation rule rather
than any specific gain as the source.

\texttt{clustered} - Clustered allocation baseline (Cluster-LUCB)
To isolate the impact of our priority-weighted allocation rule, Phase~2 evaluates against a classical ``clustered'' baseline adapted from the dueling-bandit literature (Cluster-LUCB). This approach sequentially reduces the $\mathcal{O}(N^2)$ candidate pairs by focusing exploration strictly on the active decision boundaries of the top-$k$ set. First, the $N$ entities are grouped into $K$ clusters via complete-linkage agglomerative clustering on their MRF parameter vectors, utilizing the same $\ell_1$ distance metric defined for Layer~0 (\S\ref{supp:layer0_lipschitz}). Within each cluster, a current ``champion'' entity is identified based on empirical Copeland scores. 

The algorithm then applies the LUCB (Lower Upper Confidence Bound) selection rule exclusively over this reduced set of champions. Specifically, it identifies the champion currently ranked at the boundary of the selection set (rank $k$, denoted $h$) and the highest-ranked non-selected champion (rank $k+1$, denoted $l$). The algorithm selects the next feature by solving
\begin{equation}
  f^* \;=\; \argmax_{f \in \mathrm{supp}(D_{hl})}
  \; g_{hl}(f),
  \label{eq:clustered-alloc}
\end{equation}
where $h$ is the champion currently ranked $k$th (lowest winner),
$l$ is the champion ranked $(k{+}1)$th (highest loser),
$D_{hl} = S_h - S_l$ is their diff-MRF, and
$g_{hl}(f)$ is the discriminative gain (\S\ref{sec:gain}) for that
single pair. Contrast with the priority-weighted rule
(Eq.~\ref{eq:feature_selection}), which relies on dueling bandit concept and sums gains across all
active pairs weighted by boundary proximity.

\subsection{Chaos diagnostic full protocol}
\label{supp:chaos}

The chaos diagnostic identifies patients (or, in synthetic
experiments, MRFs) for which the observation order has a
disproportionate effect on the final decision, in the sense that
multiple seeds produce disagreeing trajectories.

\paragraph{Definition.}
For a fixed patient and fixed gain/allocation method, we run the
algorithm under few seeds. Define:
\begin{itemize}
  \item \emph{Sign-flip count}: the number of rounds within a seed where
    the running decision $\sign(V^{(t)})$ flips, averaged across seeds.
    This tracks instability of the decision (which side of zero) as
    evidence accumulates, independent of magnitude.
  \item \emph{Final magnitude}: $|V^{(M)}|$, the decisiveness of the
    final answer at full observation.
  \item \emph{Chaotic-patient threshold}: a patient is classified as
    chaotic if its sign-flip count is in the top quartile \emph{and} its
    final magnitude $|V^{(M)}|$ is above the across-patient median. The
    magnitude condition is essential: it excludes genuinely ambiguous
    patients (unstable trajectory but weak final signal regardless of
    observation order), retaining only the chaotic-but-decisive cases
    where acquisition order matters even though a confident answer
    exists.
\end{itemize}

\paragraph{Use.}
When chaos stratification is invoked in the main results, the
algorithm's relative performance on chaotic patients is compared
separately from non-chaotic patients. The diagnostic is a contribution
in itself: it predicts, in advance of running the full algorithm,
which patients benefit most from inference-based methods over the
deterministic Wald baseline.

\subsection{LLM Planning Prompts}
\label{llm-planning-prompts}



\begin{promptbox}[System prompt: binary AFA --- potentials (condensed)]
You are a clinician-scientist and IBD translational
expert running active feature acquisition (AFA) on a
single transcriptomically-profiled patient.
Classify the patient as \texttt{control} vs.\
\texttt{\{condition\}} over a sequential experiment:
each turn, choose which single gene to measure nextand
re-emit a calibrated probability.
\medskip
\textbf{Knowledge source:} Two curated reference tables
for \texttt{\{condition\}} only (no control potentials).
\emph{Unary} $\theta\!\in\!\{-1,0,+1\}$: expected
direction per gene; observation matching $\theta$'s
sign is evidence \emph{for} the condition.
\emph{Pairwise} $w\!\in\!\{-1,0,+1\}$: expected
co-deviation of gene pairs; $w\!=\!+1$ same direction,
$w\!=\!-1$ opposite.
\medskip
\textbf{Per-turn output:} patient\_id,
next\_gene\_to\_measure (from unobserved list),
top\_k\_labels (length~1: \texttt{control} or
\texttt{\{condition\}}), lit\_prob (float $\in[0,1]$;
${\ge}0.5 \to$ condition, ${<}0.5 \to$ control),
supporting\_text ($\leq 600$ chars; reference observed
values and relevant potentials).
\textbf{Output:} Valid JSON only, no text outside schema.
\textbf{User prompt:} shared AFA template
(Box~\ref{box:afa-user}).
\end{promptbox}


\begin{promptbox}[System prompt: binary AFA --- literature (condensed)]
You are a clinician-scientist and IBD translational
expert running active feature acquisition (AFA) on a
single transcriptomically-profiled patient.
Classify the patient as \texttt{control} vs.\
\texttt{\{condition\}} over a sequential experiment:
each turn, choose which single gene to measure next and
re-emit a calibrated probability.
\medskip
\textbf{Knowledge source:} Retrieved abstracts for
\texttt{\{condition\}}; control has no abstracts and is
interpreted as absence of the condition signal.
Citations in supporting\_text must be prefixed with a
disease-context tag
(\texttt{[UC]}, \texttt{[DSS-colitis]}, \ldots);
do not fabricate citations.
\medskip
\textbf{Per-turn output:} patient\_id,
next\_gene\_to\_measure (from unobserved list),
top\_k\_labels (length~1: \texttt{control} or
\texttt{\{condition\}}), lit\_prob (float $\in[0,1]$;
${\ge}0.5 \to$ condition, ${<}0.5 \to$ control),
supporting\_text ($\leq 600$ chars; may reference
gene values and/or abstracts).
\textbf{Output:} Valid JSON only, no text outside schema.
\textbf{User prompt:} shared AFA template
(Box~\ref{box:afa-user}).
\end{promptbox}



\begin{promptbox}[System prompt: multiclass AFA --- literature (condensed)]
You are a clinician-scientist and IBD translational
expert running active feature acquisition (AFA) on a
single transcriptomically-profiled patient.
Plan a sequential experiment: each turn, choose which
single gene to measure next ($+1$/$-1$/$0$ relative to
healthy baseline) and return your current best top-$k$
phenotype ranking from $m$ candidates.
\medskip
\textbf{Knowledge source:} Retrieved abstracts
(\{n\_abstracts\} papers) covering the candidate
IBD phenotypes.
\medskip
\textbf{Per-turn output:} patient\_id,
next\_gene\_to\_measure (from unobserved list),
top\_k\_labels (exactly or up to $k$ labels from the
candidate list, ranked most$\to$least likely).
Optionally \texttt{done} (boolean; true only when
highly confident further measurements would not change
the prediction).
Every label must be verbatim from the candidate list;
no duplicates, no prose outside the JSON object.
\textbf{Output:} Valid JSON only, no text outside schema.
\textbf{User prompt:} shared AFA template
(Box~\ref{box:afa-user}).
\end{promptbox}


\begin{promptbox}[System prompt: multiclass AFA --- potentials (condensed)]
You are a clinician-scientist and IBD translational
expert running active feature acquisition (AFA) on a
single transcriptomically-profiled patient.
Plan a sequential experiment: each turn, choose which
single gene to measure next ($+1$/$-1$/$0$ relative to
healthy baseline) and return your current best top-$k$
phenotype ranking from $m$ candidates.
\medskip
\textbf{Knowledge source:} Two curated reference tables,
grouped by disease.
\emph{Unary} $\theta\!\in\!\{-1,0,+1\}$: expected
direction per gene; observation matching $\theta$'s
sign is evidence \emph{for} that phenotype.
\emph{Pairwise} $w\!\in\!\{-1,0,+1\}$: expected
co-deviation of gene pairs; $w\!=\!+1$ same direction,
$w\!=\!-1$ opposite.
\medskip
\textbf{Per-turn output:} patient\_id,
next\_gene\_to\_measure (from unobserved list),
top\_k\_labels (exactly or up to $k$ labels from the
candidate list, ranked most$\to$least likely).
Optionally \texttt{done} (boolean; true only when
highly confident further measurements would not change
the prediction).
Every label must be verbatim from the candidate list;
no duplicates, no prose outside the JSON object.
\textbf{Output:} Valid JSON only, no text outside schema.
\textbf{User prompt:} shared AFA template
(Box~\ref{box:afa-user}).
\end{promptbox}

\begin{promptbox}[User prompt template: AFA (shared across
  all binary and multiclass variants, issued per turn)]\label{box:afa-user}
Patient: \{patient\_id\} \quad
Turn: \{t\}/\{T\} \\
Observed genes (\{n\_obs\}/\{n\_total\}):
per-gene $\pm1$/$0$ values in observation order. \\
Unobserved genes (\{n\_unobs\}/\{n\_total\}):
comma-separated, alphabetical; pick exactly one. \\
Prior turns (optional, trajectory mode only):
turn~$n$: picked gene (observed$\,=\,v$);
top\_k$\,=\,[\ldots]$. \\
Return only JSON.
\end{promptbox}

\end{document}